\documentclass{josis}
\usepackage{hyperref}
\usepackage[hyphenbreaks]{breakurl}
\usepackage{booktabs}
\usepackage{stmaryrd}
\usepackage[T1]{fontenc}
\usepackage{cite}

\usepackage{algorithm}
\usepackage{algpseudocode}

\usepackage[table]{xcolor}
\usepackage{amssymb,amsmath}

\usepackage{graphicx}
\usepackage{makecell}
\usepackage{rotating}
\usepackage{multirow}
\usepackage{booktabs}
\usepackage{color}
\usepackage{lscape}
\usepackage{adjustbox}
\usepackage{pdfpages}
\usepackage{arydshln}
\usepackage{subcaption}
\usepackage{epstopdf}
\DeclareGraphicsExtensions{.pdf,.png}

\josisdetails{%
   number=30, year=2024, firstpage=xx, lastpage=yy, 
  doi={10.5311/JOSIS.YYYY.II.NNN},
   received={May 8, 2023}, 
   returned={Jun 24, 2023},
   revised={Aug 1, 2024},
   accepted={Oct 21, 2024}}

\makeatletter
\let\UrlSpecialsOld\UrlSpecials
\def\UrlSpecials{\UrlSpecialsOld\do\/{\Url@slash}\do\_{\Url@underscore}}%
\def\Url@slash{\@ifnextchar/{\kern-.11em\mathchar47\kern-.2em}%
    {\kern-.0em\mathchar47\kern-.08em\penalty\UrlBigBreakPenalty}}
\def\Url@underscore{\nfss@text{\leavevmode \kern.06em\vbox{\hrule\@width.3em}}}
\makeatother

\hypersetup{
colorlinks=true,
linkcolor=black,
citecolor=black,
urlcolor=black
} 

\runningauthor{\begin{minipage}{.9\textwidth}\centering Kateryna Lutsai, Christoph H. Lampert\end{minipage}}
\runningtitle{GEOLOCATION PREDICTION OF TWEETS USING TRANSFORMERS}

\begin{document}

\title{Predicting the Geolocation of Tweets Using transformer models on Customized Data}

% Insert your manuscipts authors, affiliations, and addresses
\author{Kateryna Lutsai (1)}
\author{Christoph H. Lampert (2)}

\affil {National Technical University of Ukraine “Igor Sikorsky Kyiv Polytechnic Institute”, Ukraine (1)}
\affil {Institute of Science and Technology Austria, Austria (2)}

\maketitle

\keywords{geolocation prediction, transformers, twitter dataset, machine learning, regression task, Gaussian mixture model, multitask learning}

\begin{abstract}
This research is aimed to solve the tweet/user geolocation prediction task and provide a flexible methodology for the geo-tagging of textual big data. The suggested approach implements neural networks for natural language processing (NLP) to estimate the location as coordinate pairs (longitude, latitude) and two-dimensional Gaussian Mixture Models (GMMs). The scope of proposed models has been finetuned on a Twitter dataset using pretrained Bidirectional Encoder Representations from Transformers (BERT) as base models. Performance metrics show a median error of fewer than 30 km on a worldwide-level, and fewer than 15 km on the US-level datasets for the models trained and evaluated on text features of tweets' content and metadata context. Our source code and data are available at \href{https://github.com/K4TEL/geo-twitter.git}{\nolinkurl{github.com/geo-twitter}}
\end{abstract}

\section{Introduction}\label{sec:introdiction}

Much research in recent years was dedicated to processing big data of short text corpora, such as social media posts, to extract geolocation. Location analysis provides data for personalization, making it possible to understand how social media users feel about a particular topic or issue in a specific location. Consequently, it supports social science in identifying patterns of social dynamics in specific areas or regions. It includes public health-related issues, such as vaccination or the impacts of pandemics \cite{wakamiya2018twitter}, and possible insights into the demographic characteristics of a candidate's supporters \cite{arafat2020demographic} that are valuable during Presidential elections \cite{yaqub2020location}. Besides that location analysis could be useful for governmental purposes in terms of natural disasters and crisis management, since it can improve response times, and help to allocate resources better. Furthermore, in a variety of business settings, including retail, real estate, and marketing \cite{kinsella2011m}, geolocation information provides a better understanding of people's opinions about a particular brand, product, or service in a specific location. 

Twitter, being a widely used online social network, accumulates a large volume of diverse data at a high velocity, this includes short and disordered tweets, a vast network of users and rich contextual information for both users and tweets. This data serves as input for studying three common geolocation problems such as user home location, tweet location and mentioned location prediction \cite{zheng2018survey}. Although, much research dedicated to predicting users' home location utilizes the user network of connections (e. g. followings, replies, mentions) to achieve higher accuracy \cite{ZHOU20221} \cite{zhong2020multiple} \cite{zheng2020social}, this work focuses primarily on textual data analysis. Therefore, only tweet content (the text itself) and tweet context (tweet and user metadata) are adopted to solve the problem of tweet location prediction. 

Although the user level geolocation is certainly effective for some applications, message level geolocation supports fine-grained analyses. Thus, grouped by user sets of per-tweet predictions in the form of two-dimensional distributions were leveraged for solving the task of the user's home location prediction. This methodology of using predicted from tweets GMMs for approximating user location can be found in Appendix \ref{sec:per-user-predict}.

The solving of the tweet location prediction problem requires the labeled location coordinates which are commonly extracted from the tweet's metadata. In terms of Twitter, only 1–2\% of the tweets are geo-tagged with longitude-latitude coordinates \cite{priedhorsky2014inferring} which is by default followed by the accurate place description (e. g. country, city, place name). Therefore, there are two types of location definition: a pair of numerical coordinates and a textual class label—that imply the type of a task to be either a numerical regression or a textual label classification. The latter is commonly applied for relatively small-scale problems such as location prediction on a city level \cite{li2022transformer}. 

However, this study focuses on the country and global worldwide-level scale problems, thus the former method using longitude-latitude coordinates as a labeled location was chosen as more appropriate. Since Twitter geolocation data is stored in a standard format of a coordinate system called the World Geodetic System 1984 (WGS84), the same geodetic datum was utilized to represent the tweets' labeled location. Although the WGS84 is a continuous, global reference system, when data is displayed on a two-dimensional map, such as a simple Plate carrée projection, where numerical values of longitude and latitude stay in strict ranges of 360 and 180 degrees respectively. 

The decision to solve the numerical regression rather than the label classification task using BERT-based models was also supported by the experiment results presented in Scherrer's work \cite{scherrer2021social} showing that regression outperformed classification in the geospatial accuracy (mean and median error distance similar to SAE described in Section \ref{sec:spat-metrics}) metrics for the limited Twitter datasets of BosnianCroatian-Montenegrin-Serbian and Deutch languages. Detailed descriptions of the proposed regression wrapper layers, outputs post-processing, and loss computation procedures can be found in Section \ref{sec:methodology}.

\section{Related works}\label{sec:related-works}

There are many research projects focusing on information retrieval from plain text to estimate geolocation by the application of neural networks. In terms of the previously mentioned tasks associated with Twitter data, this Section covers works related to the tweet location and the user's home location prediction. Some solutions are purely text-based and some have a hybrid approach employing the user's network of friends and mentions in addition to NLP.

According to \cite{zheng2018survey}, the granularity of user's home location prediction task has three levels: administrative regions, geographical grids, and geographical coordinates. As for the separate tweet locations, coordinates or point-of-interests (POIs) are used instead of administrative regions and grids. 

The most straightforward approach would be to process the natural languages present in social media posts. There are multiple projects processing documents or the summary of user profile tweets to predict the author geolocation \cite{roller2012supervised} \cite{wing2014hierarchical} \cite{hulden2015kernel}. These works apply geographical grids dividing the Earth's globe surface into different-sized regions such that highly populated areas are split into smaller pieces while more sparse population areas end up as large grid cells. Early efforts were mainly focused on mining indicative information from users’ posting content relying on location indicative words (LIWs) that can link users to their home locations, based on various NLP techniques (e.g., topic models and statistic models) \cite{ZHOU20221}. For example, Rahimi extracts bag-of-words features from user posts \cite{rahimi2017neural}; Wing and Baldridge estimate the word distributions for different regions \cite{wing2011simple}. Other word-centric works \cite{schulz2013multi} \cite{miyazaki2018twitter} \cite{zhang2014geocoding} also focus on filtering LIWs from the text and using the gazetteers including not only city/country names but also dialect terms to resolve the geographical indexes. While such methodologies suppose the predefined set of LIWs and their geographical coordinates, Rahimi \cite{rahimi2017continuous} proposes a neural network encoding of such words and phrases to the continuous two-dimensional space. In this light, we aimed to develop another semi-supervised method based on the state-of-the-art models for NLP with no need for LIWs datasets or geographic gazetteers.

Another source of location-indicative information is based on the user interconnection graphs, also referenced as Twitter network in Twitter-related studies. \cite{zheng2018survey}. When tasked with the prediction of users' home locations through an analysis of the Twitter network, we can assume that the user’s home is close to their friends’ home locations. Likewise, mentions and conversations allude to a closer relationship or shared interests between the two users, also referenced as social closeness. 

Furthermore, in the realm of social science, the principle of homophily \cite{mcpherson2001birds} postulates that individuals who share similarities are more inclined to establish connections at a higher rate compared to those with dissimilar attributes. Research into Twitter network utilization reveals that, in addition to mutual friendships established through the bidirectional following, users' activities such as mentions and active conversations with one another are indicative of their geographic proximity in terms of home locations \cite{mcgee2011geographic}. 

Overall, studies in the previous works show that the prediction of user home locations draws equal significance from both Twitter network structures and tweet content \cite{rahimi2015exploiting} \cite{ZHOU20221} \cite{zhong2020multiple} \cite{zheng2020social}. However, it's noteworthy that only a limited number of studies \cite{cao2015inferring} \cite{sadilek2012finding} \cite{zhai2017study} leverage the Twitter network of user connections for predicting tweet locations. 

Despite short in-length textual content, a tweet is accompanied by rich context like timestamps and geo-tags associated with it, and multiple attributes of its author profile. For instance, when sending out a tweet, it is automatically accompanied by a posting timestamp. Moreover, GPS-enabled devices such as smartphones and tablets have become widespread allowing users to easily append geo-tags to their tweets, indicating their current locations \cite{zheng2018survey}. In the case of Twitter, the spatial context is represented as machine-generated geographic object descriptions with countries, cities, place types, and full names attributed by Twitter to the user-defined geolocation coordinates. 

Additionally, users can enrich their profiles with details like hometowns, timezones, personal websites, and bios. However, common methods mainly focus on content information due to the deficient data records obtained from publicly available sources of Twitter geo-tagged datasets. Since the usage of context is not well explored for tweet location prediction \cite{zheng2018survey}, there was an opportunity for data feature engineering of tweets' metadata contained in our raw datasets of archived Twitter posts.

Some researchers also define metadata features as model inputs, usually in addition to network and textual inputs, in order to predict users' home locations \cite{miura2017unifying} \cite{huang2019hierarchical}. Among utilized metadata varieties, temporal information like tweet timestamps and user-declared timezones are shown as effective in implying tweet and home locations at coarse-grained granularity  \cite{galal2016enabling}. For instance, Yuan in the work \cite{yuan2013and} managed to cover the temporal aspect and users' mobility to estimate each tweet's location source. However, the inputs of numerical type, such as timestamps and timezone are more ambiguous for the text-oriented transformer models explored in this study, therefore, time-related features were mainly ignored on account of more relevant input options. It's important to note that the user's timezone metadata is no longer used in Twitter, but would only provide insight into the geographic region of the user's home country or its capital city.

In terms of the tweet's author profile metadata, Schulz in the work \cite{schulz2013multi}, incorporates users' self-declared home locations, personal websites, and currently outdated timezone attributes, as potential tweet location indicators. For instance, the personal websites were processed to obtain the spatial indicators such as country code and IP address. While the extracted features offer valuable insights into a user's home country, they also lead to a noticeable increase in tweet processing time. Given the imperative of maintaining a swift processing pace, especially when dealing with an overall Twitter feed generation rate of 6,000 tweets per second, the research focus shifted to utilizing readily available fields of the tweet's author profile. In particular, textual fields such as users' home locations, profile bios, names, and screen names were used as model input features for the tweet location prediction.

In the realm of user home location prediction, some authors came up with probabilistic models solving the classification task on the country/city levels \cite{han2014text} \cite{mahmud2014home}. However, employing such classification-based approaches becomes impractical for tweet location prediction, given the potential presence of hundreds of thousands of Points of Interest (POIs) within a city. While classification-based methodologies are commonly employed in home location prediction, the same does not hold true for coordinate-oriented tweet location prediction. Thus, there are relatively few studies that adopt a classification-based approach to predict the location of individual tweets in the form of specific POIs, as exemplified by \cite{cao2015inferring} and \cite{hulden2015kernel}.

Importantly, Priedhorsky's methodology was aligned with the present work as it utilized GMMs for geolocation prediction instead of simple coordinates prediction \cite{priedhorsky2014inferring}. Another subsequent study, which employed a hybrid approach, also estimated geolocation through GMMs rather than coordinates \cite{bakerman2018twitter}. This study's final estimation was based on the text and network features as joint predictors. The performance metrics used in both works were adopted to evaluate the proposed probabilistic models as described in Section \ref{sec:prob-metrics}. 

Continuing the exploration of probabilistic models used in the relevant studies, Iso in the work \cite{iso2017density} proposed the convolutional mixture density network which is applied to the tweets' content to estimate the parameters of the GMM and employ the mode value of estimated distribution density as the predicted coordinates for tweets. The authors also claim that the choice of a loss function affects the model performance and describe a basic logic of the probabilistic model loss computation using negative log-likelihood as shown in Eq. \eqref{eq:nllh}. Similarly, Li in the work \cite{li2019geoattn} utilizes the negative log-likelihood of the GMM over all training examples. Notably, the weights of the Gaussian components were obtained directly from the model's attention scores. In contrast, we propose the prediction of all GMM parameters (weights, covariance) by the model itself rather than the application of available attention weights. 

Most of the previously mentioned works use various approaches based on neural networks for NLP, but none of them has applied transformer models for the estimation of geolocation. The relatively new BERT model (released in 2018 \cite{devlin2018bert}) is aimed at classification tasks and some works have already used it for country/city names predictions. 

The base BERT model is intended to be finetuned on a downstream task which, in this case, was a regression to geographical coordinates (and GMM parameters such as weights and covariance) output as a type of sentence classification task. In general, BERT models have been pretrained using Masked Language Modeling (MLM) to improve the model's understanding of bidirectional sentence representations, along with Next Sentence Prediction (NSP) to capture the inter-sentence relationships. Since BERT’s layers are hierarchical, early BERT layers learn more generic linguistic patterns, such as differences between multiple languages and their general sentence structures (in the case of the BERT multilingual model). While the later layers learn more task-specific patterns, in this case—the associations between points on the world map and specific terms referred to as LIWs, as well as text constructions and linguistic patterns commonly used in certain geographical areas.

For instance, Villegas in the work \cite{villegas2020point} predicts place type or point of interest (POI) using lowercase English BERT. Furthermore, Li in the work \cite{li2022transformer} focused on city-scale prediction for the Twitter datasets of Melbourne, and Singapore. However, the model presented by Li took the text and metadata input and gave the result in a form of probabilities for the set of POI classes.

In the study conducted by Simanjuntak \cite{simanjuntak2022we}, the task of predicting Twitter users' home location was explored using Long-Short Term Memory (LSTM) and BERT models. The authors used a dataset in the Indonesian language containing text content and user metadata context of tweets. Given that the base model has a limited input capacity of 512 tokens (words), it is not feasible to process large text corpora that comprise all user tweets.

Such constraints were challenging for classifying a user's home location, as concatenating all of the user's tweets into a single text would result in an input exceeding the limit. Authors state that the majority vote approach for per-tweet predictions in 42\% cases resulted in misclassification of user home location. Thus only the task of tweet geolocation prediction could be efficiently solved using the BERT-base models. In contrast, this work focuses on processing smaller (less than 300 words) text samples and aggregating a set of probabilistic predictions in a form of GMMs to estimate the set of most probable user's home location points as described in Section \ref{sec:per-user-predict}. 

In another work, Scherrer uses BERT models for both regression and classification tasks on limited text datasets of BosnianCroatian-Montenegrin-Serbian and Deutch languages \cite{scherrer2021social}. Evaluation of their regression model on the bound to German-speaking Switzerland dataset shows 21.20 and 30.60 km median and mean distance errors. The authors state that using a smaller tokenization vocabulary and converting the geolocation task to a classification task, in general, yielded worse results. They also conclude that hyperparameter tuning did not yield any consistent improvements, and simply selecting the optimal epoch number on development data showed to be the best approach to the problem of geolocation prediction.

Moreover, the results presented by \cite{scherrer2021social} indicate that regression outperformed classification in geolocation prediction using modified BERT models. Therefore, this work aims to estimate the geolocation of single tweets in the form of multiple possible location points represented by geographical coordinates (longitude, latitude, weight) or two-dimensional GMMs (longitude, latitude, weight, covariance). As for the task of user's home location prediction, the output of probabilistic models involving the covariance parameter is reduced to a set of weighted coordinate pairs (longitude, latitude, weight). 

In this work, the geographic granularity level was defined as a worldwide area covering the whole scope of countries and languages present in the Twitter feed. However, the results of Scherrer's study \cite{scherrer2021social} demonstrated that language-specific BERT models clearly outperforms their multilingual counterpart on most of the used datasets. In light of these findings, the country-level model for the English and French-speaking countries was finetuned on a limited subset of French, Canadian, and British tweets as described in Section \ref{sec:ablation}.

Nevertheless, the global dataset was multilingual so the corresponding BERT base model ('bert-base-multilingual-cased') was chosen, which resulted in higher error distance metrics for the per-tweet and per-user geolocation prediction, as expected. A critical challenge addressed is the representation of multiple languages, since the ability to process multiple languages is a key factor in dividing text inputs into geospatial regions. To tackle this issue, a multilingual BERT model, pretrained on the 104 largest Wikipedia languages, was utilized as a starting point for finetuning on a global dataset with geolocation labels.

\section{Materials and Methods}\label{sec:methodology}

The main goal of this work was location prediction in a form of geographic points and two-dimensional distributions based on short texts processed by the modified BERT models. The common approach to solving the regression task for BERT models is adding the dense linear layer on top of the classification output tokens. To account for the multitask learning, individual wrapper layers were used for key and minor textual features. Architecture modifications for the best of proposed models involved only the scope of wrapper layers implementing linear regression to the output of parameters defining a GMM (means coordinates, weights, and covariance parameters).

\begin{figure}[!htb]
\centering
     \includegraphics[width=1.0\textwidth]{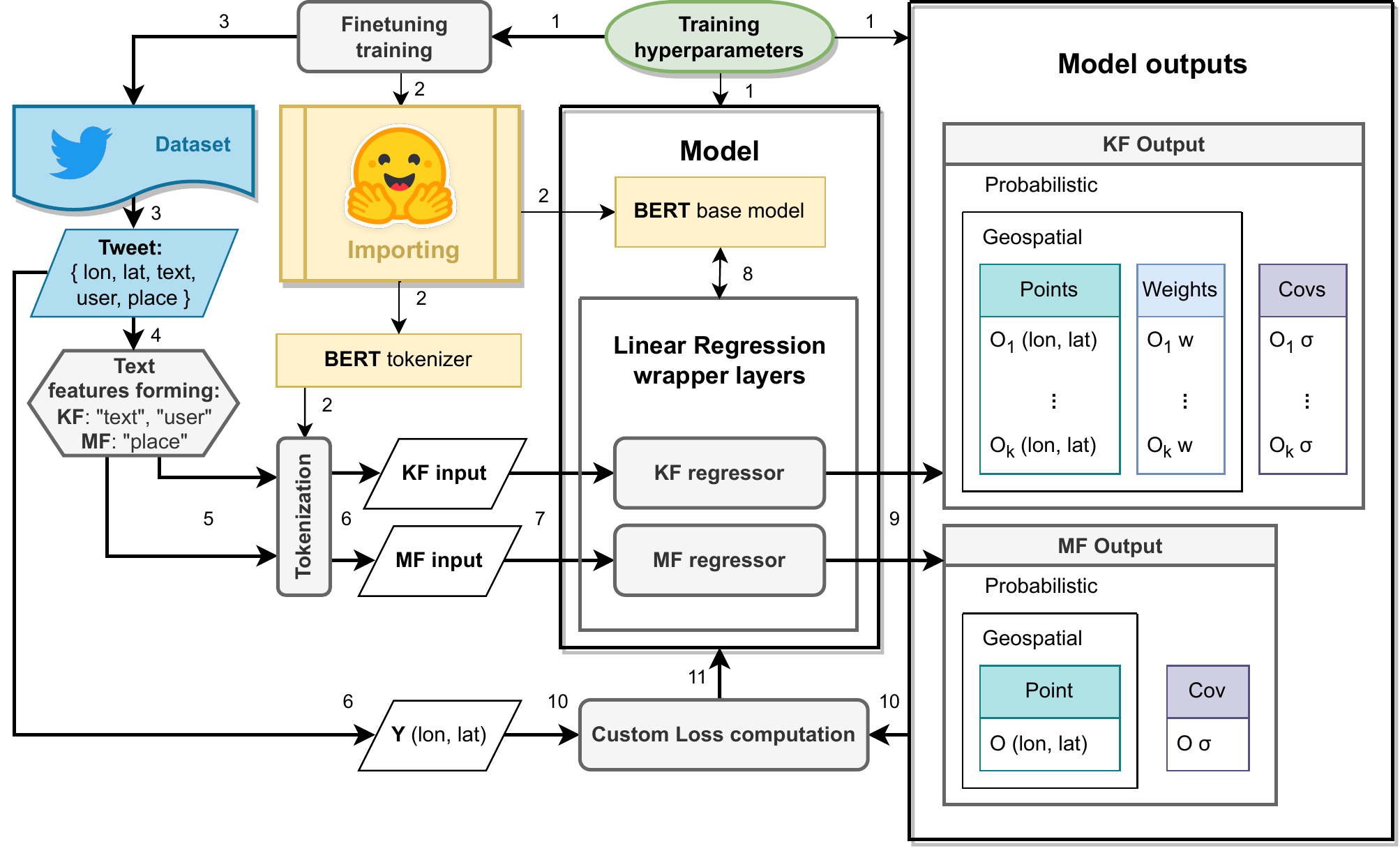}
      \caption{Model finetuning procedure flowchart for the best of the proposed models which has the Probabilistic Multiple Prediction Outcomes (PMOP) output type and individual wrapper layers for the necessary Key Feature (KF) and optional Minor Feature (MF) \\
      Step 1: parameterization of the model's architecture, outputs, and finetuning; \\ 
      Step 2: importing default tokenizer and base BERT \cite{devlin2018bert} model; \\
      Step 3: reading dataset files from the Twitter archive into a virtual sheet; \\
      Step 4: forming inputs \textit{NON-GEO} as KF and \textit{GEO-ONLY} as MF according to Section \ref{sec:data-preprocess}; \\
      Step 5: tokenization of the text inputs through the standard BERT tokenizer; \\
      Step 6: forming of the model data loaders by selection of batch subsets from data; \\
      Step 7: loading input features to the model, each to separate wrapper layer; \\
      Step 8: text processing by the local base BERT model before applying wrapper layers; \\
      Step 9: returning defined in size model outputs according to Table \ref{tab:model-type-outputs}; \\
      Step 10: comparing processed predictions and labels according to Section \ref{sec:loss-functions}; \\
      Step 11: backpropagation of the total per-batch loss to the local base BERT version.}
       \label{fig:model-train}
\end{figure}

Considering that the two-dimensional distributions were used as the output format of the probabilistic models, this work proposes the custom procedures of loss function computation for the four model types different in output formats. The training procedure for the best of the proposed models is demonstrated on the flowchart diagram in Figure \ref{fig:model-train}. In terms of the input data used for model fine-tuning, this study put into service only textual information such as the tweet content (raw user's text) and tweet context (metadata like information from user's profile and geo-tag textual descriptions). 

The regression layer was used to convert the vector of final hidden states made up of 768 floats to the specified number of numerical outputs. According to the task of geolocation prediction, the essential number of outputs was defined as 2 standing for the geographic point (longitude, latitude) in the WGS84 coordinate system. However, the best of the proposed model had an output of 20 numerical values utilized as parameters for the definition of predicted GMM. Note that the models predicting a single coordinate pair could be trained using the standard Mean Squared Error (MSE) loss function. However, more complex output forms such as GMMs required a custom loss function described in Section \ref{sec:architecture}. The intention was to test whether the Probabilistic Single/Multiple Outcome Prediction (PSOP/PMOP) models outperform the straightforward approach of Geospatial Single/Multiple Outcome Prediction (GSOP/GMOP) models.

The hyper-parameters tuning stage has covered such parameters as the type of the scheduler, minimal and maximal learning rates for the chosen scheduler, and the number of epochs. For the current task, the cosine type was chosen from the set of linear, cyclic, step, and plateau schedulers. Experiments have shown that the optimal learning rate reduction range starts at 1e-5 and ends at 1e-6 at the end of the last training epoch. The entering number of epochs was reduced from 5 to 3 for the dataset consisting of 2.7 million training samples, which are grouped into batches of 10 to 16 tweets in the model data loader. This decision was made as a result of the observed lack of significant error distance reduction after the third epoch for both training and test (development set) metrics. 
After each epoch, the models were evaluated on the development set, which consists of 300,000 samples. 
%By default, the development set of  was evaluated without a gradient propagation at the end of each epoch. 
The described hyper-parameters were used for all proposed models, with variations only occurring in the loss function type, the number of outcomes (prediction points), covariance type, and the combination of text features.

\subsection{Data preprocessing}\label{sec:data-preprocess}

The datasets used for finetuning the models consisted of tweets collected between the years 2020–2022 which have defined 'coordinates' and 'place' objects in the tweet JSON. The datasets have undergone several preprocessing stages prior to being uploaded into the model, including text strings filtering, rearrangement of the columns to form text features, tokenization, and split into tensor data loaders of preset batch size. The model was operating solely with input IDs and attention masks representing text features of the dataset and its numerical labels which corresponded to geolocation coordinates.

Generally, only 5\% of all tweets in our Twitter archive database collected from 2020 to 2022 have geolocation coordinates, such that the number of available geo-tagged samples was approximately 150 million. It has been found that up to 20\% of the geo-tagged tweets in the collected datasets were posted by bot users, which are automated accounts e.g., newspapers, weather forecasts, airplane trackers, etc. While the finetuning datasets contained tweets of both real and bot users, the evaluation datasets were filtered out based on the assumption that real users don't post more than 20 messages per day. 

The training data we collected for this study includes a comprehensive set of tweets and their metadata (author profiles, location geo-tags) serving as inputs, and labeled location coordinates playing role of the models's 'ground truth' through the training and evaluation. During the preprocessing stage, the original tweet objects were condensed into the text content ('text'), relevant metadata context ('user' and 'place'), and geolocation coordinates ('lon', 'lat') representing the labeled location. 

Worth mentioning that the word-centric techniques, which use geocoders to recognize the referred place based on the geographic gazetteers \cite{schulz2013multi}. The gazetteer typically includes entries for cities, towns, villages, landmarks, natural features, and other geographic entities, along with information such as their location, coordinates, elevation, etc. The challenge in this work was to find a substitute for the accurate knowledge base of geographical references (e.g., gazetteer terms). 

Fortunately, Twitter metadata stored in the 'place' field of geo-tagged tweets provides automatically generated information about the country, country code, place type, location name, and location full name. Since tweet JSON root-level objects “place” and “user” had multiple fields containing potentially relevant context data, some of them were used in forming the input features as shown in Table \ref{tab:feature-fields}. Despite the fact that only 26\% of users provide location information in their profiles according to \cite{cheng2010you} and that user-generated data can be noisy, the proposed multitask learning procedures ought to enhance the geospatial accuracy. 

\begin{table}[!htb]
\small
\centering
\begin{tabular}{ccccc|l|l}
\multicolumn{5}{c|}{\textbf{Feature name}} & \multicolumn{1}{l|}{\multirow{2}{*}{\begin{tabular}[l]{@{}l@{}}\textbf{Dataset}\\\textbf{columns}\end{tabular}}} & \multicolumn{1}{c}{\multirow{2}{*}{\begin{tabular}[l]{@{}l@{}}\textbf{JSON}\\\textbf{fields}\end{tabular}}} \\
ALL & TEXT-ONLY & USER-ONLY & GEO-ONLY & NON-GEO & \multicolumn{1}{c|}{} & \multicolumn{1}{c}{} \\ 
\hline
\textbf{+} & \textbf{+} &  &  & \textbf{+} & text & text \\ 
\hline
\textbf{+} &  & \textbf{+} &  & \textbf{+} & user & \begin{tabular}[c]{@{}l@{}}location, \\ description, \\ name,\\ screen\_name\end{tabular} \\ 
\hline
\textbf{+} &  &  & \textbf{+} &  & place & \begin{tabular}[c]{@{}l@{}}country, \\ place\_type,\\ location, \\ name, \\ full\_name\end{tabular}
\end{tabular}
\caption{Text feature contents formed from the dataset of parsed tweet JSON fields}
\label{tab:feature-fields}
\end{table}

Since typical model input shouldn't include 'place' context which refers exclusively to geo-tagged tweets, multitask learning has been implemented in order to both preserve the input format and utilize 'place' metadata during training. This approach implies separate wrapper layers for each of the textual input features and a per-batch loss common for all features during finetuning. The main input feature for the best of the proposed models included the tweet text Content combined with context obtained from the author profile (username, description, and location), while the supplemental input feature was dedicated solely to the Context data obtained from the 'place' field. 

Considering that evaluation ought to be performed on the typical model input ('text' and 'user'), models trained with the described Key and Minor Feature setup were expected to show lower spatial errors in comparison to the models trained without the minor feature or the models trained on a combination of all data in a single text feature ('text', 'user', and 'place'). Moreover, the tweet metadata providing context for the Content of the tweet was expected to improve overall model accuracy by comparison with the 'TO' (\textit{TEXT-ONLY}) model trained only on the 'text' contents without metadata. More details on the comparison of merged and split input features can be found in Section \ref{sec:discussion} and specifically in Table \ref{tab:worldwide-metrics} containing all performance measurements. 

Our research primarily focused on using text information to predict tweets and users' home locations, as it is easily accessible from tweet contents and user profile data. Since all tweets are linked to their authors, user profiles were utilized as a major part of the model input sequences. Although profile background images could provide additional insights, incorporating them would require a more resource-intensive hybrid text+image model, which was beyond the scope of our work. Furthermore, the utilization of portrait images would rise more serious privacy issue of sensitive data.

One of the limitations encountered is the unavailability of appropriate datasets that include profile background images. The Twitter datasets commonly used for location prediction do not have these images, and the Twitter streaming API does not provide immediate access to them, requiring manual retrieval of each image individually. This process would introduce significant overhead, especially for a continuously running system.

Considering the challenges of computational complexity and data availability, we chose to concentrate solely on user profile textual information (screen name, name, description, and location of the Twitter user profile), as shown in Table \ref{tab:feature-fields}. This allowed us to effectively achieve our research objectives of predicting geolocation for the arbitrary tweets within our available resources and datasets. Worth mentioning, that the user's home location wasn't retrieved from the profile but rather sampled from the user's tweets subset during evaluation. 

The labeled location for this study was defined as the pair of longitude-latitude coordinates associated with each geo-tagged tweet. By definition, there are two JSON children entities that describe the location information of a tweet: 'coordinates' and 'place'. The former is present only when the exact location was assigned by the user, therefore data labels in a longitude-latitude format have been collected by parsing only the 'coordinates' object of tweets. 

In contrast to the tweet numerical geo-tag, the 'place' field provides an accurate automatically generated by Twitter textual description of the location, such as country-city naming, which was utilized as a supplementary input during the finetuning of the best models. As shown in Table \ref{tab:feature-fields}, the fields of the 'place' object utilized in forming the training dataset included country, place type, name, location, and full name of the geo-tagged location. 

The decision to split the training input into Key and Minor Features was dictated by the assumption about the model's attention shift. The model trained on the combination of all input features was expected to learn to pay attention only to the geo-tags. To avoid memorizing of positions of the geo-tag tokens in the input sequence during training, the randomization of words could not be applied due to the semantics distortion. 

The comparison of 'A' (\textit{ALL}) and 'NG+GO' (\textit{NON-GEO}+\textit{GEO-ONLY}) models was intended to prove that the identical model setups trained on merged and split input features would differ in accuracy. In case the separation of \textit{GEO-ONLY} from the always available \textit{NON-GEO} enhanced the \textit{NON-GEO} accuracy, it would support the idea of applying multitask learning techniques in this work despite the slower in total processing of multiple inputs and outputs. 

There was a limited number of text feature combinations explored in this study as described in Table \ref{tab:approach-feature}. Note that each training sample had separate tokenized sequences for all input features assigned to it in the model data loader, such that per-batch loss was computed as a linear combination of per-feature errors. The visualization of the multitask loss computation procedure can be found in Figure \ref{fig:total-loss} alongside the scheme for a single feature (Key only) model of K outcomes in Figure \ref{fig:mop-loss}. 

\begin{table}[!htb]
\small
\centering
\begin{tabular}{c|c|c|c}
\multirow{2}{*}{\textbf{Data type}} & \multicolumn{2}{c|}{\textbf{Training Features }} & \multirow{2}{*}{\textbf{Accuracy }} \\
 & \textbf{Key} & \textbf{Minor} &  \\ 
\hline
Content & TEXT-ONLY & \multirow{3}{*}{-} & low \\ 
\cline{1-2}\cline{4-4}
\multirow{4}{*}{Content, Context} & NON-GEO &  & medium \\ 
\cline{2-2}\cline{4-4}
 & ALL &  & medium \\ 
\cline{2-4}
 & TEXT-ONLY & USER-ONLY, GEO-ONLY & low \\ 
\cline{2-4}
 & \textbf{NON-GEO} & \textbf{GEO-ONLY} & \textbf{high}
\end{tabular}
\caption{Text features combinations used in text-based (Content) and hybrid (Content and Context) approaches.}
\label{tab:approach-feature}
\end{table}

The models were evaluated on diverse arbitrary tweets imitating a common situation when the 'place' field is unavailable and only the key features of always present “text” and “user” inputs are used for prediction. In general, each tweet always contains “text” and “user” (\textit{NON-GEO}) used as the Key training Feature, and as an input during the evaluation of development and validation datasets. 

In practice, each model had a necessary Key Feature (KF) and an optional Minor Feature (MF) wrapper layers which were used according to the input type during the finetuning stage. Utilizing the 'place' field as MF and user-generated text ('text' and 'user') as KF was crucial for predicting geolocation coordinates accurately. The 'place' field provided valuable complementary information, enhancing context understanding and disambiguating similar locations, while tweet content offered relevant cues and linguistic patterns, improving prediction precision. 

The described combination of \textit{NON-GEO} KF and \textit{GEO-ONLY} MF created a robust and flexible model, capable of handling diverse tweets when the 'place' field is unreliable or unavailable. Therefore, the resulting trained system possesses the capability to effectively analyze and predict the labeled location coordinates based on any arbitrary input of the tweet-like length.

The final stage of preparing data for model uploading involved converting the dataset's text features into a data loader of numeric tensors with corresponding IDs, attention masks, and labels. This process started with filtering out URLs and messy punctuation from the text to eliminate irrelevant information. The purified text is then encoded using the default BERT tokenizer, transformed into data loaders with a specified batch size, and made ready for use by the model. Each sample in the batch of data loader represented a single tweet input encoded to IDs and attention masks of its text features, and labeled by its geographic coordinate pair. In practice, the number of words in the tokenized text corpus remained within the limit of 300, which was suitable for the base BERT model's input size of 512 tokens (words).

\subsection{Model architecture}\label{sec:architecture}

In this work, it is noted that the BERT multilingual base model was tasked with performing regression to predict numerical values like coordinate values, weights, and covariance matrix parameters. This downstream task is similar to text classification and requires modification of the final layer of the model. The base BERT model has a hidden size of 768, which is also the size of the hidden-state token for sequence classification. 

\begin{table}[!htb]
\small
\centering
\begin{tabular}{l|c|ccc|cc|c}
\multirow{2}{*}{\textbf{Type}} & \multirow{2}{*}{\begin{tabular}[c]{@{}c@{}}\textbf{Prediction}\\\textbf{Outcome}\end{tabular}} & \multicolumn{3}{c|}{\textbf{Key Feature}} & \multicolumn{2}{c|}{\textbf{Minor Features}} & \multirow{2}{*}{\textbf{Code}} \\
 &  & \multicolumn{1}{c}{Point} & \multicolumn{1}{c}{Weight} & \multicolumn{1}{c|}{Cov} & \multicolumn{1}{c}{Point} & \multicolumn{1}{c|}{Cov} &  \\ 
\hline
\multirow{2}{*}{Spat} & 1 & \textbf{2} & - & - & \multirow{2}{*}{\textbf{2}} & \multirow{2}{*}{-} & GSOP \\ 
\cline{2-5}\cline{8-8}
 & M$>$1 & \textbf{M*2} & \textbf{M} & - &  &  & GMOP \\ 
\hline
\multirow{2}{*}{Prob} & 1 & \textbf{2} & - & \textbf{1} & \multirow{2}{*}{\textbf{2}} & \multirow{2}{*}{\textbf{1}} & PSOP \\ 
\cline{2-5}\cline{8-8}
 & M$>$1 & \textbf{M*2} & \textbf{M} & \textbf{M} &  &  & PMOP
\end{tabular}

\caption{Wrapper layer types by the number of outputs grouped by Spatial (Spat) and Probabilistic (Prob) output forms; Prediction Outcomes determines the number of predicted geographic points; Key Feature and Minor Features stand for the wrapper layers of the proposed models; \\
Code column shows the abbreviations of different model configs. \\
Point refers to coordinate pair $\mathbf{\widehat{Y}}$; Weight is $\widehat{w}$ converted to the GMM weights $W$; and Cov corresponds to $\widehat{c}$ converted to the $\sigma_{\widehat{c}}$ parameter of the spherical covariance matrix $\mathbf{\Sigma}$}
\label{tab:model-type-outputs}
\end{table}

The proposed wrapper layer operates using only the BERT model pooler output, which consists of processed classification tokens, each representing a separate tweet in the batch. The wrapper layer implements a common linear regression logic, transforming the vector of size 768 to an output vector of a specified size. The exact number of outputs depends on the type of model being used. Furthermore, all models could be grouped into 4 types by the difference in their output structure as shown in Table \ref{tab:model-type-outputs}.

The wrapper layer implemented linear regression with a dynamic number of outputs which depended on the feature type (Key or Minor), model type (Geospatial or Probabilistic), and the number of prediction outcomes (Single or Multiple). In the simplest case of the Single Outcome Prediction (SOP) key difference between Geospatial and Probabilistic models was the form of output which could be a two-dimensional point or a Bivariate Normal Distribution.

\[\mathbf{\widehat{Y}_{spat}}=(\widehat{y}_{lon}, \widehat{y}_{lat}); \quad \widehat{y}_{lon}\in \mathbb{R}; \quad \widehat{y}_{lat}\in \mathbb{R}\]
\[\widehat{Y}_{prob}=N(\boldsymbol{\widehat{\mu}},\mathbf{\Sigma}); \quad \boldsymbol{\widehat{\mu}} = \mathbf{\widehat{Y}_{spat}}; \quad \mathbf{\Sigma}=\begin{bmatrix}
\sigma_{\widehat{c}} & 0\\
0 & \sigma_{\widehat{c}}
\end{bmatrix}; \quad \sigma_{\widehat{c}}>0\]

The probabilistic output of the model included not only the spatial component of the predicted coordinate pair $\boldsymbol{\widehat{\mu}}$ but also a measure of the model's confidence in its prediction. The Gaussian covariance matrix, which represents the uncertainty of the model, was of spherical type and could be defined by a single positive nonzero numerical value $\sigma$. To ensure that $\sigma_{\widehat{c}}$ remains positive, the SoftPlus function described in Eq. \eqref{eq:sp} could be applied to the output variable $\widehat{c}$. 

\begin{equation}
    \sigma_{\widehat{c}} = \log(1 + e^{\widehat{c}}); \quad \widehat{c} \in \mathbb{R}; \quad \sigma_{\widehat{c}} > 0 \tag{SoftPlus} \label{eq:sp}
\end{equation}

However, a lower bound for $\sigma_{\widehat{c}}$ was established at $\frac{1}{2\pi}$ to preserve values of the predicted Gaussian's Probability Density Function described in Eq. \eqref{eq:pdf} in the range of [0, 1].

\begin{equation}
    \sigma_{\widehat{c}} = \log(1 + e^{\widehat{c}})+\frac{1}{2 \pi}; \quad \widehat{c} \in \mathbb{R} ; \quad \sigma_{\widehat{c}} \in (\frac{1}{2\pi}, +\infty) \tag{LBSP} \label{eq:lbsp}
\end{equation}

The selection of the spherical covariance matrix was based on empirical evidence revealing lower geospatial errors compared to models utilizing diagonal, full, and tied covariance matrices. While more complex matrices such as diagonal and full provide more freedom in the shape of the distribution, they necessitate the production of more outputs. Opting for the minimal number of values to determine the shape of the Gaussian distribution eliminated the risk of the model prioritizing optimization of the probabilistic loss component described in Eq. \eqref{eq:nllh} over the spatial component of the loss described in Eq. \eqref{eq:sed}. 

\begin{figure}[!htb]
  \centering
  \begin{minipage}{0.5\textwidth}
    \includegraphics[width=\linewidth]{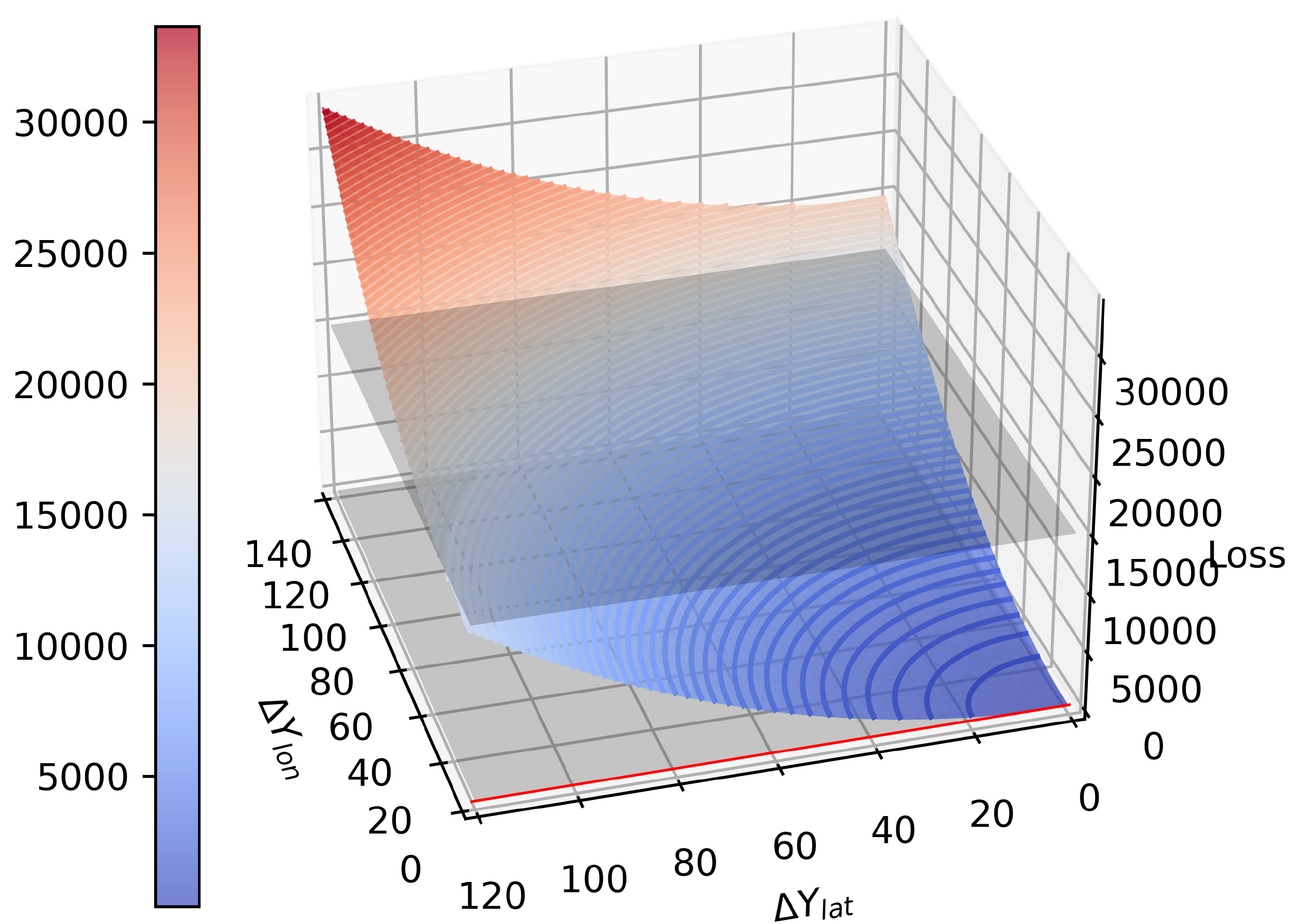}
      \caption{Squared Euclidean Distance \eqref{eq:sed} function surface on the axes of $\Delta Y_{lon}$ and $\Delta Y_{lat}$ as the error distances per longitude and latitude axes; upper horizontal gray surface indicates the empirical maximum of $L_{spat}$; red line indicates the strict minimum of 0 implied by the nature of Eq. \eqref{eq:sed}.}
       \label{fig:loss-graph-spat}
  \end{minipage}\hfill
  \begin{minipage}{0.48\textwidth}
    \includegraphics[width=\linewidth]{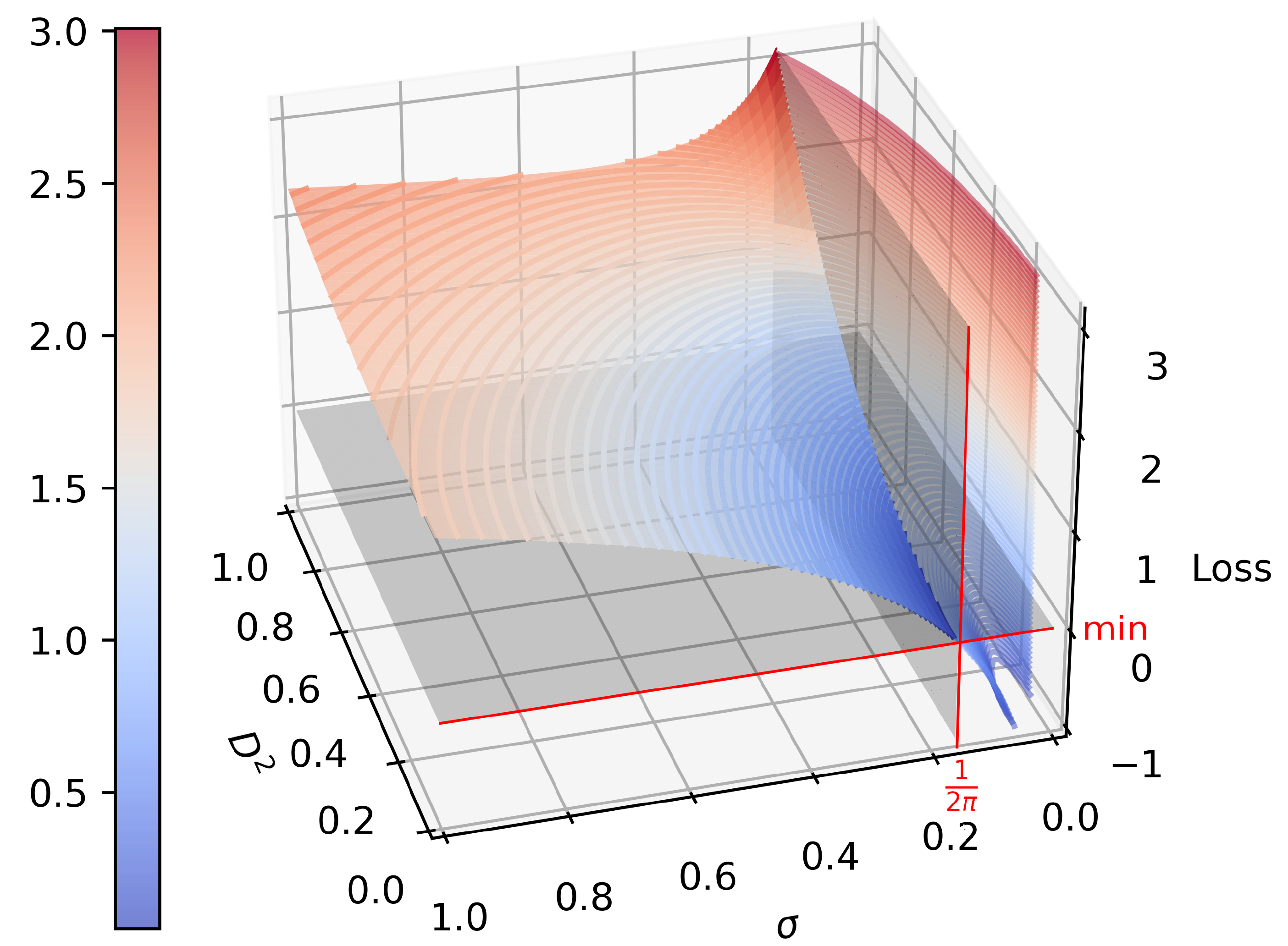}
      \caption{Negative Log-LikeliHood \eqref{eq:nllh} function surface on the axes of $D^2$ as the error distance and $\sigma_{\widehat{c}}$ as the uncertainty in $\widehat{\mu}$ of the Gaussian; red lines and gray surfaces indicate the reduction of $L_{prob}$ domain as a result of Eq. \eqref{eq:lbsp} application.}
       \label{fig:loss-graph-prob}
  \end{minipage}
\end{figure}

Moreover, the revised lower bound $\frac{1}{2\pi}$ for $\sigma_{\widehat{c}}$ curtailed the sharpness of the Gaussian peaks and ensured that their height never exceeds 1, thus avoiding negative values in the probabilistic loss as show in Figure \ref{fig:loss-graph-prob}. As a result of applying the lower-bounded SoftPlus described in Eq. \eqref{eq:lbsp} on the outputs associated with the covariance parameter, both $L_{spat}$ and $L_{prob}$ remained in the positive domain and were approaching 0 during the finetuning. Hence it eliminated the loss momentum difference and made it possible to include both geospatial and probabilistic errors as tantamount loss components during the total loss computation. 

\begin{figure}[!htb]
\centering
     \includegraphics[width=1.0\textwidth]{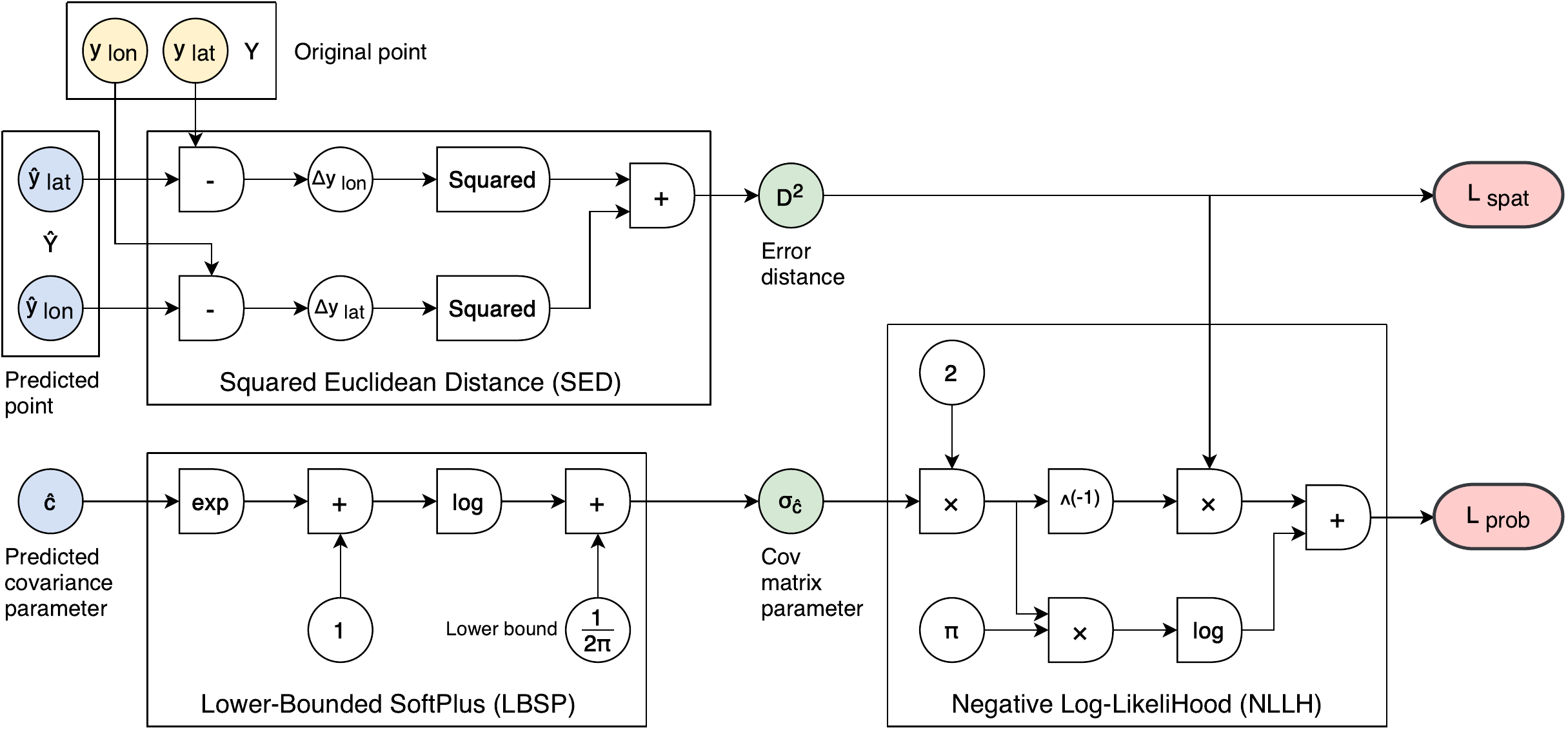}
      \caption{Single Outcome Prediction (SOP) model loss functions computational graph including visualization of Squared Euclidean Distance \eqref{eq:sed}, Lower-Bounded SoftPlus \eqref{eq:lbsp}, and Negative Log-LikeliHood \eqref{eq:nllh} components.}
       \label{fig:sop-loss}
\end{figure}

To account for the variations in the output variables, each model type required an individual method for computing its loss. Such that models of the Geospatial type were utilizing solely $L_{spat}$, while models of the Probabilistic type were utilizing a combination of both $L_{spat}$ and $L_{prob}$. In terms of KF output evaluation, loss computation procedures for the SOP-type models are visualized in Figure \ref{fig:sop-loss}, while computational graphs for the MOP-type models and multitask learning procedures can be found in Appendix \ref{sec:loss-functions}. 

The spatial loss for SOP-type models was computed as the squared Euclidean distance between the true location specified by the user and the location predicted by the model on the two-dimensional Plate carrée projection of the worldwide map. Note that the $L_{spat}$ by its nature always remains greater than or equal to zero as shown in Figure \ref{fig:loss-graph-spat}. 

\begin{equation}
    L_{GSOP} = (\mathbf{Y}-\mathbf{\widehat{Y}})^2 = (y_{lon}-\widehat{y}_{lon})^2+(y_{lat}-\widehat{y}_{lat})^2 = D^2; \quad D \geq 0 \tag{SED} \label{eq:sed}
\end{equation}

The probabilistic component of PSOP models loss was computed as the negative log-likelihood for the original point $\mathbf{Y}$ to fit in the predicted Gaussian distribution $N(\boldsymbol{\widehat{\mu}}, \mathbf{\Sigma})$ as described in Eq. \eqref{eq:nllh}.

\begin{equation}
    PDF = N(\mathbf{Y} \mid \boldsymbol{\widehat{\mu}}, \mathbf{\Sigma})=\frac{e^{-\frac{D^2}{2\sigma}}}{2\pi\sigma}; \quad PDF\in[0,1] \tag{PDF} \label{eq:pdf}
\end{equation}

\begin{equation}
    L_{PSOP}= - \log(PDF) = \frac{D^2}{2\sigma} + \log(2\pi\sigma); \quad L_{PSOP}\geq 0 \tag{NLLH} \label{eq:nllh}
\end{equation}

Importantly, the definition of the Gaussian covariance matrix implies $\sigma$ to be greater than 0 which could be achieved by the application of Eq. \eqref{eq:sp} with a lower bound of 0. However, assuming that the error distance is approaching 0, model uncertainty in a predicted point would approach 0 as well. As a consequence, Eq. \eqref{eq:pdf} would exceed 1 resulting in negative values of Eq. \eqref{eq:nllh}:

\[\lim_{(\sigma, D) \to (0^+, 0^+)} PDF = + \infty; \quad \lim_{(\sigma, D) \to (0^+, 0^+)} - \log(PDF) = - \infty\]

This study suggests the application of Eq. \eqref{eq:lbsp} to limit the covariance parameter as shown in Figure \ref{fig:loss-graph-prob}. Hence, when the measure of uncertainty $\sigma$ is approaching its minimum at $\frac{1}{2\pi}$, $L_{prob}$ relies mainly on its spatial component $D^2$:

\[\lim_{\sigma \to \frac{1}{2\pi}} PDF = e^{-\pi D^2}; \quad \lim_{\sigma \to \frac{1}{2\pi}} L_{PSOP} = \pi D^2;\]

In the case of MOP-type models, the output included a weight $\widehat{w}_{i}$ for each of $M$ outcomes which indicated its significance among other outcomes. To ensure that all predicted weights sum up to 1, the SoftMax function described in Eq. \eqref{eq:sm} was applied to $\widehat{w}_{i}$.        

\begin{equation}
    W_i=\frac{e^{\widehat{w}_{i}}}{\sum_{j=1}^{M}e^{\widehat{w}_{j}}}; \quad \widehat{w}\in \mathbb{R}; \quad \sum_{i=1}^{M}W_i = 1; \quad W_i \in [0, 1] \tag{SoftMax} \label{eq:sm}
\end{equation}

Therefore, the total of MOP geospatial and probabilistic loss was computed as the weighted linear combination of all $M$ outcomes errors described in Eq. \eqref{eq:wlc}.

\begin{equation}
    L_{GMOP}=\sum_{i=1}^{M}W_{i}D_{i}^{2}; \quad L_{PMOP}=\sum_{i=1}^{M}W_{i}(\frac{D_i^2}{2\sigma_{i}} + \log(2\pi\sigma_{i})) \tag{WLC} \label{eq:wlc}
\end{equation}

The total loss per text feature $f$ had an essential geospatial component, such that $L_{f}$ was equal to the spatial loss $L_{spat}$, and optionally the probabilistic loss $L_{prob}$ added to it. The per-feature loss for the probabilistic models was computed in three ways: average or sum of $L_{spat}$ and $L_{prob}$, or $L_{prob}$ with no regard for $L_{spat}$. The first option has slightly outperformed the second option and has significantly outperformed the last option, thus per-feature loss was computed as an average of two components in the case of probabilistic models.  

Finally, the total loss as an average of all per-feature loss values was computed to handle multiple textual features of a single tweet. 

In practice, the final step was computing the mean of total loss per tweet in a single data loader tensor (batch) to back-propagate a float value representing the total per-batch loss. The experiments also revealed a computational time growth of 12\% in the case of probabilistic models compared to the geospatial analogs. Our equipment (NVIDIA GeForce GTX 1080 Ti) was able to process geospatial loss calculations of 16,7 tweets per second (t/s) in comparison to the 14,9 t/s of the combined geospatial and probabilistic loss computations needed in the case of probabilistic models.

To sum up, Section \ref{sec:architecture} covers a scope of all models reviewed in Table \ref{tab:model-type-outputs}. According to the experimental results in Table \ref{tab:worldwide-metrics}, the most accurate model configuration was the Probabilistic Multiple Outcomes Prediction (PMOP) model of 5 outcomes utilizing \textit{NON-GEO} as the Key Feature and \textit{GEO-ONLY} as the Minor Feature, which was finetuned on a worldwide dataset of tweets using custom loss computation procedures described in Appendix \ref{sec:loss-functions}.  

\subsection{Results}\label{sec:results}

In this section, the proposed models are evaluated by geospatial and probabilistic performance metrics in global (Table \ref{tab:worldwide-metrics}) and country-level (Table \ref{tab:ablation-metrics}) granularity Twitter datasets. Importantly, the final evaluation was performed on the datasets filtered from bot users (posting more than 20 messages per day) and users utilized in the train datasets to ensure that the evaluation was unbiased and provided an accurate assessment of the models' generalization capabilities.  

The visualization of the post-processed prediction example for the best of the proposed models is shown below in Figure \ref{fig:prediction-example}. The predicted set of weighted two-dimensional points and two-dimensional distribution with selected peaks were visualized for the Geospatial and Probabilistic models trained by the best approach. This example demonstrates a prediction of the randomly chosen text without a labeled location (genuine truth) for the sake of comparison between two types of models trained on the Content and Context and producing an output of either multiple geolocation points (GMOP) or united distributions of location likelihood (PMOP).  Considering the training data and procedures were the same for both models, there was a 15-degree difference between the GMOP and PMOP predictions. The indirect input without geographic references resulted in a relatively small shift of less than 2,000 km while maintaining the common guess of the US origins of the text. 

\begin{figure}[!htb]
\centering
     \includegraphics[width=1.0\textwidth]{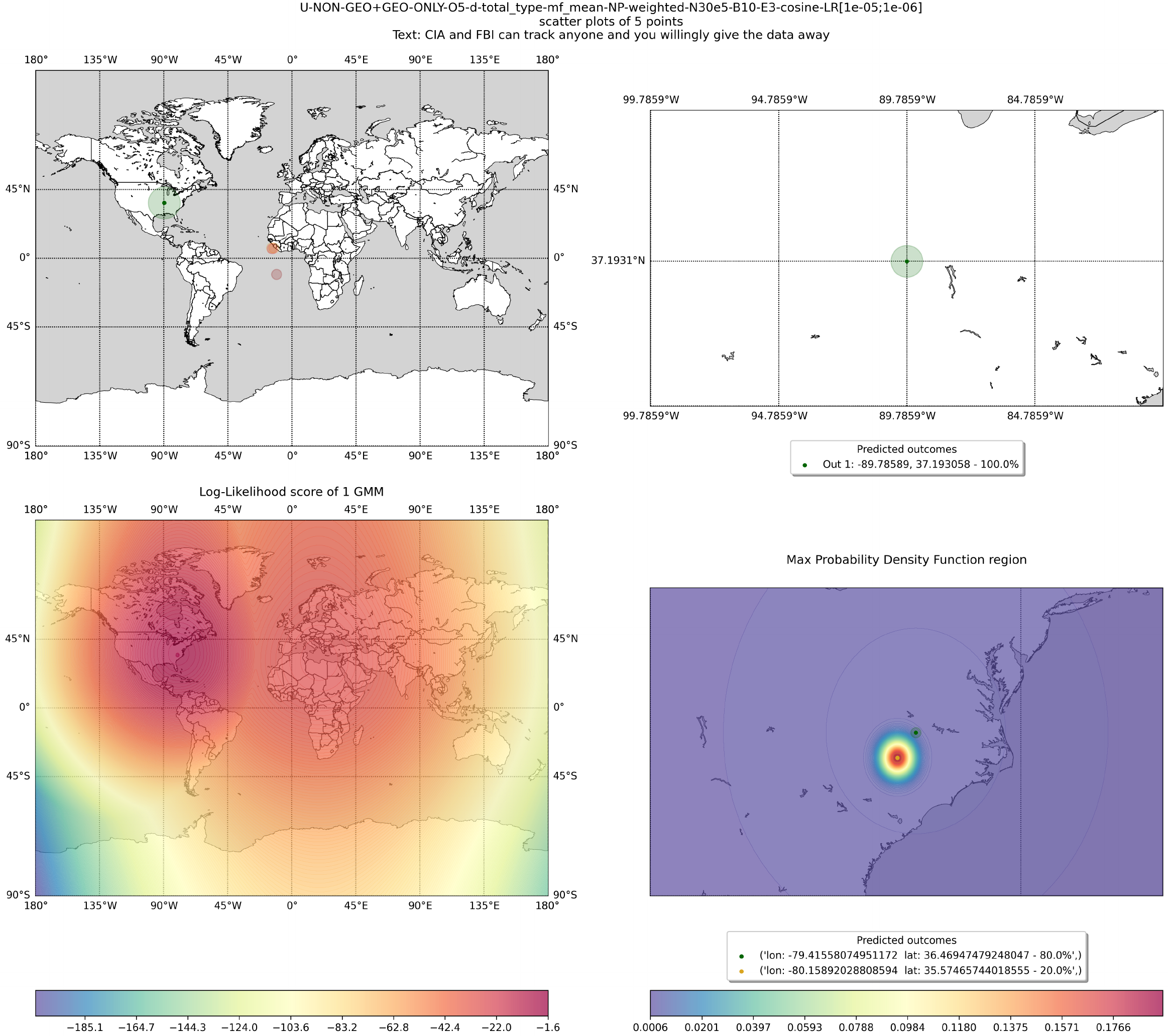}
      \caption{Prediction examples of two models, which were trained on the same  worldwide dataset with key \textit{NON-GEO} and minor \textit{GEO-ONLY} text features, for the same text: \\
      \textit{'CIA and FBI can track anyone, and you willingly give the data away'} with 5 outcomes sorted by significance (weight);  \\
      above: Geospatial GMOP points as scatter plots; \\
      below: Probabilistic PMOP Gaussian peaks as LLH and PDF plots.}
       \label{fig:prediction-example}
\end{figure}

\subsubsection{Worldwide evaluation metrics by model type}\label{sec:worldwide-metrics}

The models compared in Table \ref{tab:worldwide-metrics} were trained on the local Twitter-World dataset composed of different Train Features (TF) and tested on both \textit{TEXT-ONLY} and \textit{NON-GEO} eValuation Features (VF). The most prevalent \textit{NG+GO} training setup was using the hybrid approach with \textit{NON-GEO} as the Key and \textit{GEO-ONLY} as the Minor Features respectively. Metrics results for the MOP-type models were computed as the weighted linear combination of all outcomes, specific formulas for the MOP-type model can be found in Section \ref{sec:performance-metrics}. 

Evaluation results of the worldwide models in Table \ref{tab:worldwide-metrics} show the difference between geospatial and probabilistic models with a number of prediction outcomes ranging from 1 to 100. The development dataset consisted of 300K tweets from 143K users covering 63 languages and 226 countries. Note that both train and development datasets were the same for all models and contained tweets from both real and bot users.

\begin{table}[!htb]
\small
\centering
\begin{adjustbox}{width=1.0\textwidth}
\begin{tabular}{r|l|l|c|ccc|ccccc}
\multirow{2}{*}{\textbf{TF}} & \multirow{2}{*}{\textbf{Model}} & \multirow{2}{*}{\textbf{VF}} & \multirow{2}{*}{\textbf{OUT}} & \multicolumn{3}{c|}{\textbf{Spatial}} & \multicolumn{5}{c}{\textbf{Probabilistic}} \\
 &  &  &  & \begin{tabular}[c]{@{}c@{}}\textbf{Mean}\\\textbf{ SAE}\end{tabular} & \begin{tabular}[c]{@{}c@{}}\textbf{Med}\\\textbf{ SAE}\end{tabular} & \begin{tabular}[c]{@{}c@{}}\textbf{Acc}\\\textbf{ @161}\end{tabular} & \begin{tabular}[c]{@{}c@{}}\textbf{Mean}\\\textbf{ CAE}\end{tabular} & \begin{tabular}[c]{@{}c@{}}\textbf{Med}\\\textbf{ CAE}\end{tabular} & \begin{tabular}[c]{@{}c@{}}\textbf{Mean}\\$\mathbf{PRA_{0.95}}$\end{tabular} & \begin{tabular}[c]{@{}c@{}}\textbf{Med}\\$\mathbf{PRA_{0.95}}$\end{tabular} & $\mathbf{COV_{0.95}}$ \\ 
\hline
\multirow{18}{*}{\rotcell{\textbf{NG+GO}}} & \multirow{2}{*}{PSOP} & TO & \multirow{2}{*}{1} & 1881.2 & 153 & 50.5 & 1954 & 352.8 & 174 & 60.8 & 12.5 \\
 &  & NG &  & 568.3 & 32.1 & 78.3 & 639.2 & 70.4 & 60.9 & \textbf{3.7} & 19.6 \\ 
\cline{2-12}
 & \multirow{12}{*}{\textbf{PMOP}} & \multicolumn{1}{c|}{TO} & \multirow{2}{*}{3} & 1876.2 & 134.9 & 51.5 & 1911.8 & 219.9 & 24.1 & 16.2 & 22 \\
 &  & \multicolumn{1}{c|}{NG} & & 561.7 & 29.5 & 78.7 & 601.1 & \textbf{63.3} & \textbf{13.9} & 7.9 & \textbf{31.2} \\ 
\cline{3-12}
 &  & TO & \multirow{2}{*}{\textbf{5}} & 1845 & 135.9 & 51.5 & 1896.3 & 214.7 & 27.9 & 15.2 & 15.9 \\
 &  & \textbf{NG} &  & \textbf{551.4} & 29.4 & \textbf{79} & \textbf{600.4} & 79.1 & 15.3 & 9.6 & 23.4 \\ 
\cline{3-12}
 &  & \multicolumn{1}{c|}{TO} & \multirow{2}{*}{5u*} & 2036.7 & 234.8 & 45.1 & 2080.9 & 357.5 & 27.9 & 28.8 & 6.2 \\
 &  & \multicolumn{1}{c|}{NG} &  & 626.7 & 70 & 70.7 & 691 & 158.5 & 21.3 & 18.5 & 9.7 \\ 
\cline{3-12}
 &  & TO & \multirow{2}{*}{10} & 1845 & 143.2 & 51.1 & 1901.4 & 214.2 & 32.4 & 13.9 & 7.8 \\
 &  & NG &  & 553.2 & 33.2 & 78.9 & 599.8 & 77.9 & 15.6 & 9.4 & 11 \\ 
\cline{3-12}
 &  & TO & \multirow{2}{*}{50} & 1855.9 & 136.6 & 51.4 & 1905 & 214.7 & 27.8 & 14 & 1.3 \\
 &  & NG &  & 556.9 & 29.4 & \textbf{79} & 604.4 & 77.1 & 14.7 & 9.5 & 2.1 \\ 
\cline{3-12}
 &  & TO & \multirow{2}{*}{100} & 1882.4 & 149.8 & 50.6 & 1939 & 199.6 & 193.2 & 4.5 & 13.9 \\
 &  & NG &  & 568.9 & \textbf{28.1} & 78.6 & 618.3 & 71.1 & 15.2 & 9.1 & 1.2 \\ 
\cline{2-12}
 & \multirow{2}{*}{GSOP} & TO & \multirow{2}{*}{1} & 1872.2 & 140.3 & 51.3 & \multicolumn{5}{c}{\multirow{6}{*}{-}} \\
 &  & NG &  & 559.6 & 36.6 & 78.4 & \multicolumn{5}{c}{} \\ 
\cline{2-7}
 & \multirow{4}{*}{GMOP} & \multicolumn{1}{c|}{TO} & \multirow{2}{*}{3} & 1859.7 & 142.8 & 51.1 & \multicolumn{5}{c}{} \\
 &  & NG &  & 556.3 & 36.5 & 78.5 & \multicolumn{5}{c}{} \\ 
\cline{3-7}
 &  & TO & \multirow{2}{*}{5} & 1986.6 & 151.3 & 50.6 & \multicolumn{5}{c}{} \\
 &  & NG &  & 577.8 & 35.6 & 78.6 & \multicolumn{5}{c}{} \\ 
\hline
\multirow{2}{*}{\rotcell{A}} & \multirow{6}{*}{PMOP} & TO & \multirow{6}{*}{5} & 3203 & 585.9 & 41.2 & 3225.9 & 623.4 & 29.6 & 8.5 & 7.4 \\
 &  & NG &  & 1266.2 & 37.7 & 67 & 1292.6 & 62.6 & 14.7 & 7.5 & 11.7 \\ 
\cline{1-1}\cline{3-3}\cline{5-12}
\multirow{2}{*}{\rotcell{NG}} &  & TO &  & 1875.4 & 172.9 & 49.3 & 1927.3 & 235.2 & 24.6 & 13.8 & 7.5 \\
 &  & NG &  & 581 & 46.9 & 76.4 & 635.2 & 107 & 15.9 & 11.8 & 12.6 \\ 
\cline{1-1}\cline{3-3}\cline{5-12}
\multirow{2}{*}{\rotcell{TO}} &  & TO &  & 1547.3 & 176.5 & 48.7 & 1708.8 & 442.1 & 58.8 & 38.2 & 8.8 \\
 &  & NG &  & 782 & 87.2 & 64.9 & 913.1 & 267.1 & 36.2 & 26.4 & 11.8
\end{tabular}
\end{adjustbox}
\caption{Worldwide dataset (300,000 tweets) results of models performance metrics; \\
    TF stands for Training Feature, VF for eValuation Feature, and OUT defines number of outcomes while other model configuration parameters remain steady; \\ 
    u*—unlimited covariance output, $\sigma_{\widehat{c}}$ were obtained by the application of Eq. \eqref{eq:sp} to the covariance parameter output $\widehat{c}$ \\
    A: \textit{ALL} or 'text'+'user'+'place', NG: \textit{NON-GEO} or 'text'+'user', TO: \textit{TEXT-ONLY} or 'text', \\
    NG+GO: \textit{NON-GEO} + \textit{GEO-ONLY} or ('text'+'user')+'place' where 'text', 'user', and 'place' refer to the inputs defined in Table \ref{tab:feature-fields}.   
}
\label{tab:worldwide-metrics}
\end{table}

\begin{figure}[!htb]
  \centering
  \begin{minipage}{\textwidth}
    \centering
    \includegraphics[width=\linewidth]{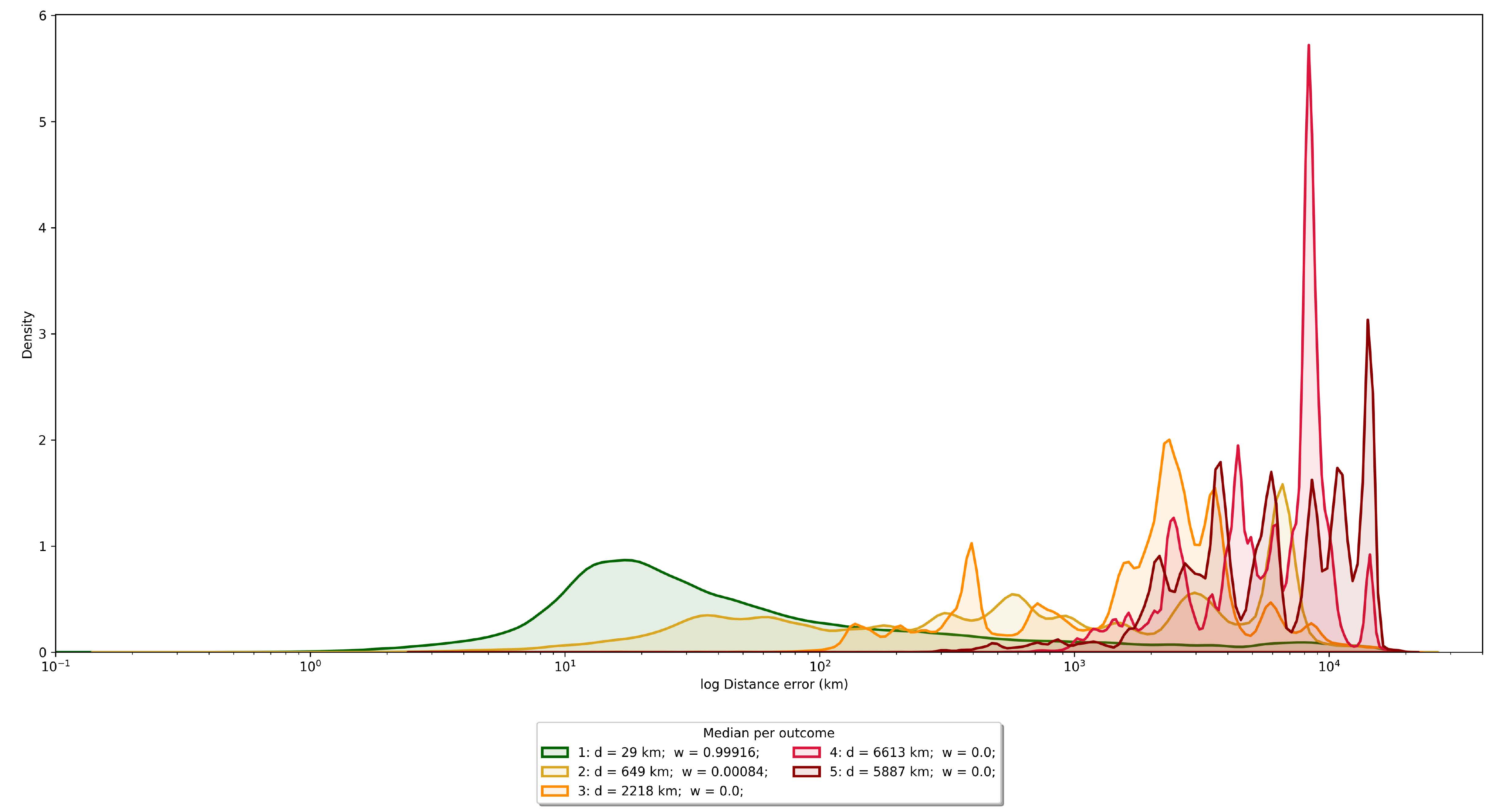}
  \end{minipage}
  \begin{minipage}{\textwidth}
    \centering
    \includegraphics[width=\linewidth]{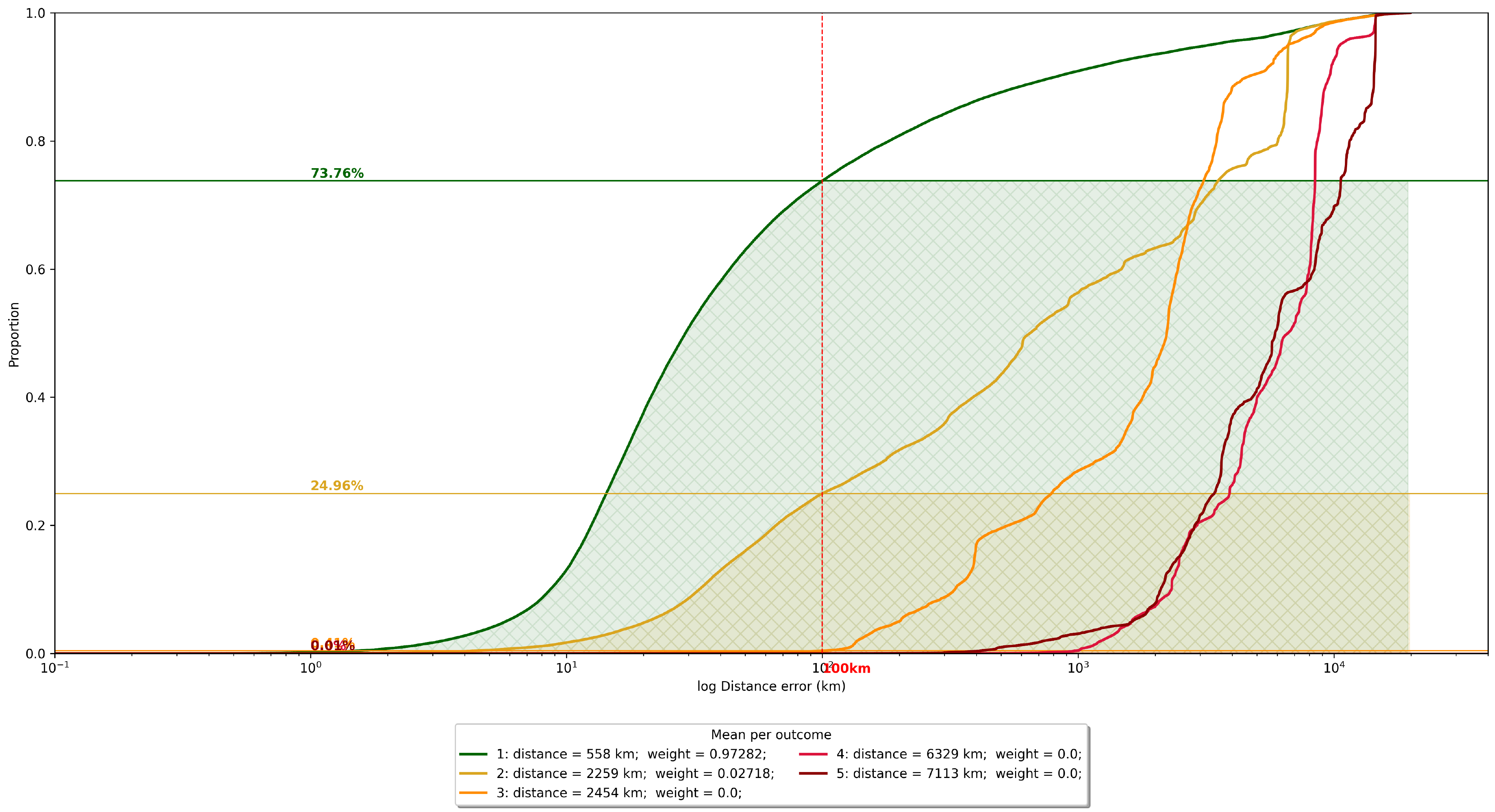}
  \end{minipage}
  \caption{Density (above) and Cumulative distribution (below) of the log error distance in km per outcome for the best model (Probabilistic, 5 outcomes, \textit{NON-GEO} + \textit{GEO-ONLY}) computed on the dataset of 100,000 worldwide tweets; median (above) and mean (below) errors and weights per outcome are noted in the legends below the graphs.}
  \label{fig:stats-worldwide}
\end{figure}

Following the best of the proposed approaches, the total per-tweet loss was composed of two features containing both spatial and probabilistic components. To account for the equivalent of both $L_{spat}$ and $L_{prob}$ loss momentums during finetuning, Eq. \eqref{eq:lbsp} set $\frac{1}{2\pi}$ as the lower bound of $\sigma_{\widehat{c}}$ covariance parameters thereby maintaining the values of Eq. \eqref{eq:nllh} in the positive domain. The performance results in Table \ref{tab:worldwide-metrics} confirm that the PMOP model of 5 outcomes noted by $u*$ (stands for the obtained from Eq. \eqref{eq:sp} unlimited $\sigma_{\widehat{c}}$) showed downscale scores in all metrics in comparison to the lower-bounded model of 5 outcomes. 

As for the Training Feature options, the last three models in Table \ref{tab:worldwide-metrics} were trained on \textit{ALL} ('text' + 'user' + 'place'), \textit{NON-GEO} ('text' + 'user'), and \textit{TEXT-ONLY} ('text') features had higher SAE scores compared to the similar PMOP model of 5 outcomes trained on the proposed combination of \textit{NON-GEO} and \textit{GEO-ONLY} features, indicating underperformance. 

Moreover, the increasing number of outcomes resulted in interchangeable spatial metrics scores, while revealing a modest growth of CAE and PRA metrics, as well as an undesirable decrease in the COV criteria. According to the analysis of the error distance per outcome depicted in Figure \ref{fig:stats-worldwide}, up to 2 outcomes were assigned significant weights and only 73.76\% predictions of the best outcome had median errors below 100 km.

The Probabilistic Single/Multiple Outcome Prediction (PSOP/PMOP) models have slightly outperformed the straightforward approach of Geospatial Single/Multiple Outcome Prediction (GSOP/GMOP) models on the median error distance metrics as shown in Table \ref{tab:worldwide-metrics}. This observation supports findings in the work \cite{li2019geoattn} that demonstrated the benefits of predicting a probability distribution rather than a single point - several candidate locations shifted the single output in different directions. Similarly, the MOP-type architecture outperformed SOP-type in both Probabilistic and Geospatial variations, though the difference was relatively low. 

\subsubsection{Ablation study metrics}\label{sec:ablation}

In contrast to the worldwide dataset metrics, the ablation study of the best model covers only three countries Great Britain, France, and Canada. The choice of these three countries was made to specifically address the challenge of handling ambiguity in geolocation prediction on multilingual territories. Canada, in particular, was included because tweets from this country can be in either English or French, which often leads to confusion for models that cannot accurately predict the probability distributions. 

\begin{table}[!htb]
\small
\centering
\begin{tabular}{l|cccc}
\multirow{2}{*}{\textbf{Stat}} & \multicolumn{4}{c}{\textbf{Country}} \\
 & \multicolumn{1}{c}{\textbf{All}} & \multicolumn{1}{c}{\textbf{Great Britain}} & \multicolumn{1}{c}{\textbf{France}} & \textbf{Canada} \\ 
\hline
Train size & 3,000,000 & 1,772,344 & 366,882 & 860,834 \\
English lang. & 2,494,749 & 1,649,065 & 53,753 & 791,931 \\
French lang. & 326,165 & 17,052 & 281,163 & 27,950 \\
Other lang. & 179,086 & 106,227 & 31,906 & 40,953 \\ 
\hline
Test size & 300,000 & 176,894 & 36,607 & 86,499 \\
English lang. & 249,750 & 164,661 & 5,424 & 79,665 \\
French lang. & 32,344 & 1702 & 25,905 & 2,737 \\
Other lang. & 17,906 & 10,531 & 5,278 & 4,097 \\ 
\hdashline
Mean SAE & 57.19 km & 31.15 km & 65.51 km & 106.93 km \\
Med SAE & 3.68 km & 3.3 km & 5.9 km & 3.94 km \\
Acc@161 & 96\% & 97\% & 92\% & 95\%
\end{tabular}
\caption{Performance results of the model trained and evaluated on the Great Britain + France + Canada dataset using the hybrid approach (Content + Context) and PMOP-type output of 5 prediction outcomes for the estimation of the tweet location.}
\label{tab:ablation-metrics}
\end{table}

The aim of the ablation study was to showcase the handling of ambiguity and evaluate the performance of the model in a limited dataset, focusing on these specific countries. The dataset composition was designed to demonstrate the impact of language distribution and the proportion of samples from each country on the model's accuracy.

Table \ref{tab:ablation-metrics} presents the statistics of the ablation study, illustrating the distribution of tweets and languages within the dataset. British tweets constituted the largest proportion, accounting for 59\% of the training and evaluation dataset, Canadian samples represented 29\%, while French tweets had the smallest portion at 12\%. Additionally, the distribution of languages showed that English tweets dominated the dataset, comprising 83\% of all tweets, while French tweets accounted for 11\%, and the remaining 6\% were in other languages. Since 75\% of the French and only 3\% of Canadian tweets were in the French language the small number of French samples was compensated by their distinctive nature. 

Therefore, the ablation study on these specific countries effectively highlighted the model's ability to handle ambiguity in language and demonstrated the influence of dataset composition on geolocation prediction accuracy. The ablation study results demonstrated that the model performed better on this limited dataset compared to the worldwide and American datasets. Notably, 96\% of the predictions were within 161 km of the ground truth locations, which supports the suggestion \cite{scherrer2021social} of training country-scale models. 

Among the selected countries, France and Great Britain, with their relatively smaller square areas, facilitated more accurate geolocation predictions. The proportion of samples per country within the dataset also influenced the model's accuracy, with British tweets outnumbering French tweets by a factor of 5, resulting in a two-fold increase in the mean and median SAE of the French data segment. Overall, Great Britain exhibited the lowest mean and median error distances, France, with the smallest amount of data, had the highest median SAE, and Canada, with the largest square area, had the highest mean SAE metrics.

\section{Discussion and Conclusion}\label{sec:discussion}

This work was aimed at the examination of the machine learning techniques for solving the short text geolocation prediction task with NLP techniques employing a scope of BERT-based neural networks. The proposed framework demonstrates simplicity in terms of the trained model application as a tool for geo-tagging textual big data of short corpora. While the process of finetuning the models involved some complexity in loss functions, it is a one-time step that requires initial time and computing resources to develop the methodology of training the most accurate models. 

The designed model architecture is adaptive and simple in structure since there is only a superficial modification of the base model's wrapper layers implementing linear regression with no changes to the base model layers. The models are flexible to any level of the geospatial granularity (worldwide, country, city, etc) by changing the finetuning dataset and language-specific pretrained base BERT model. Once the finetuning is completed, the resulting model becomes essentially plug-and-play, requiring minimal additional effort for user-specific downstream tasks.  

For instance, setup requirements for the proposed model application are the following: access to the local or remote model at \href{https://huggingface.co/k4tel/geo-bert-multilingual}{\nolinkurl{huggingface/geo-bert-multilingual}}; optionally dataset of text samples shorter than 512 tokens; and GPU or CPU node to launch the model. 

While the training routine requires access to the base model; dataset that contains labeled coordinates (lon, lat), text features for the arbitrary model input, and supplementary geo-tags of the locations; the usage of GPU is recommended due to the time and computing resources needed. 

\begin{figure}[!htb]
  \centering
  \begin{subfigure}[b]{0.49\textwidth}
    \centering
    \includegraphics[width=\linewidth]{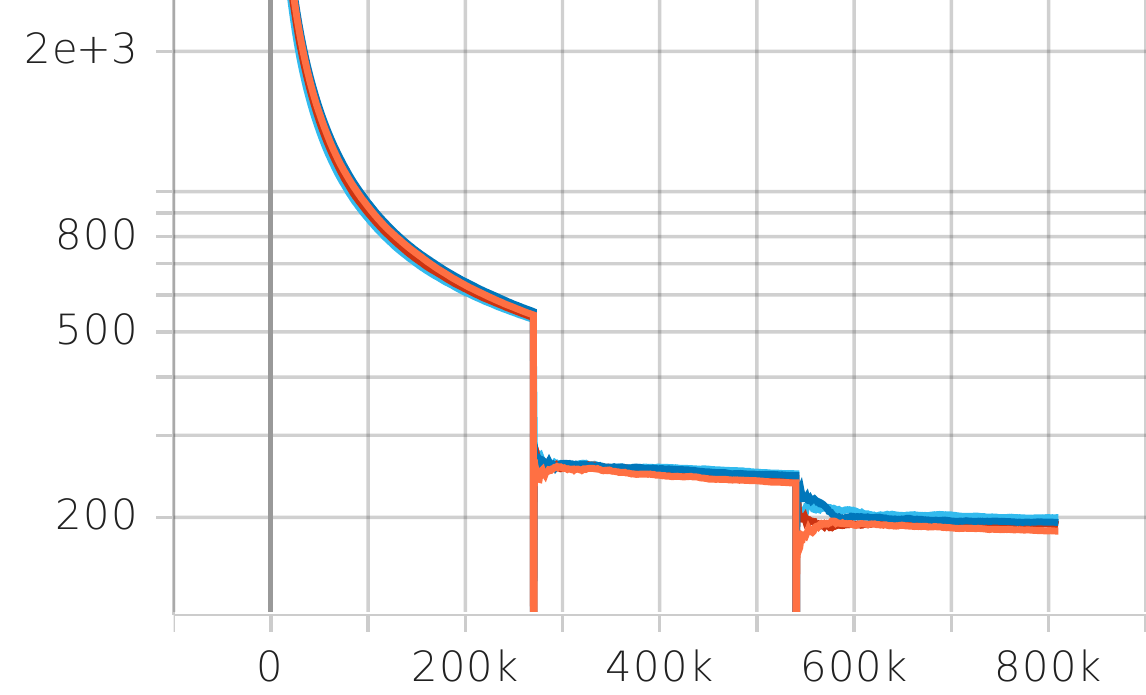}
    \caption{KF Geospatial loss computed by Eq. \eqref{eq:sed}}
  \end{subfigure}
  \hfill
  \begin{subfigure}[b]{0.49\textwidth}
    \centering
    \includegraphics[width=\linewidth]{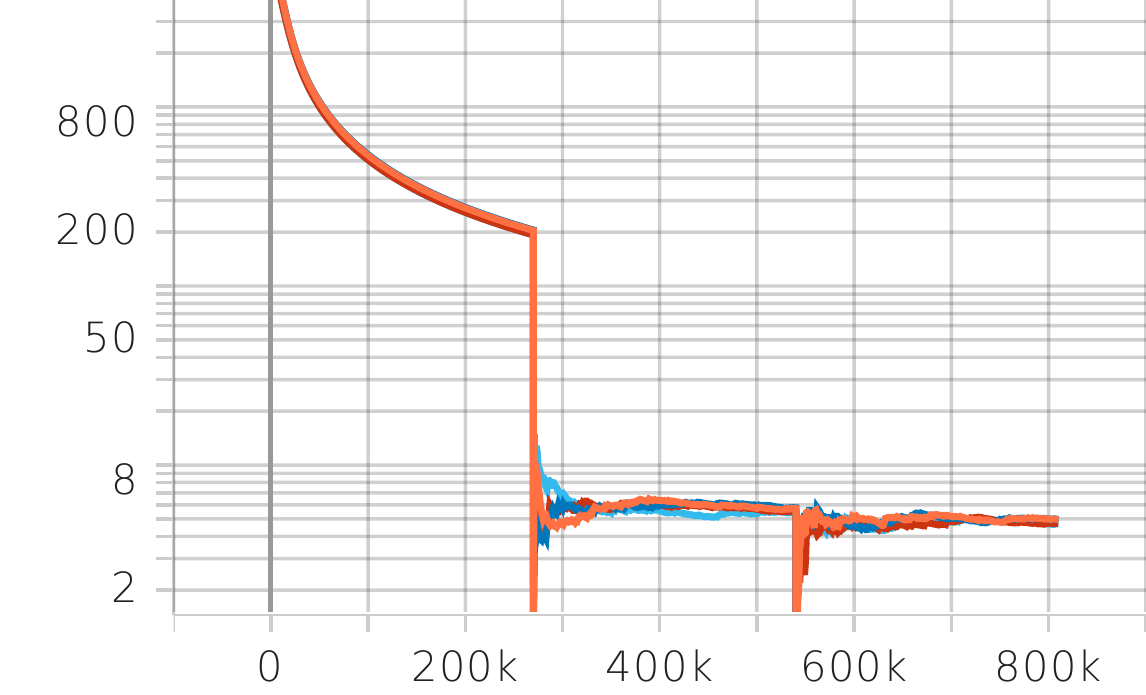}
    \caption{MF Geospatial loss computed by Eq. \eqref{eq:sed}}
  \end{subfigure}

  \medskip

  \begin{subfigure}[b]{0.49\textwidth}
    \centering
    \includegraphics[width=\linewidth]{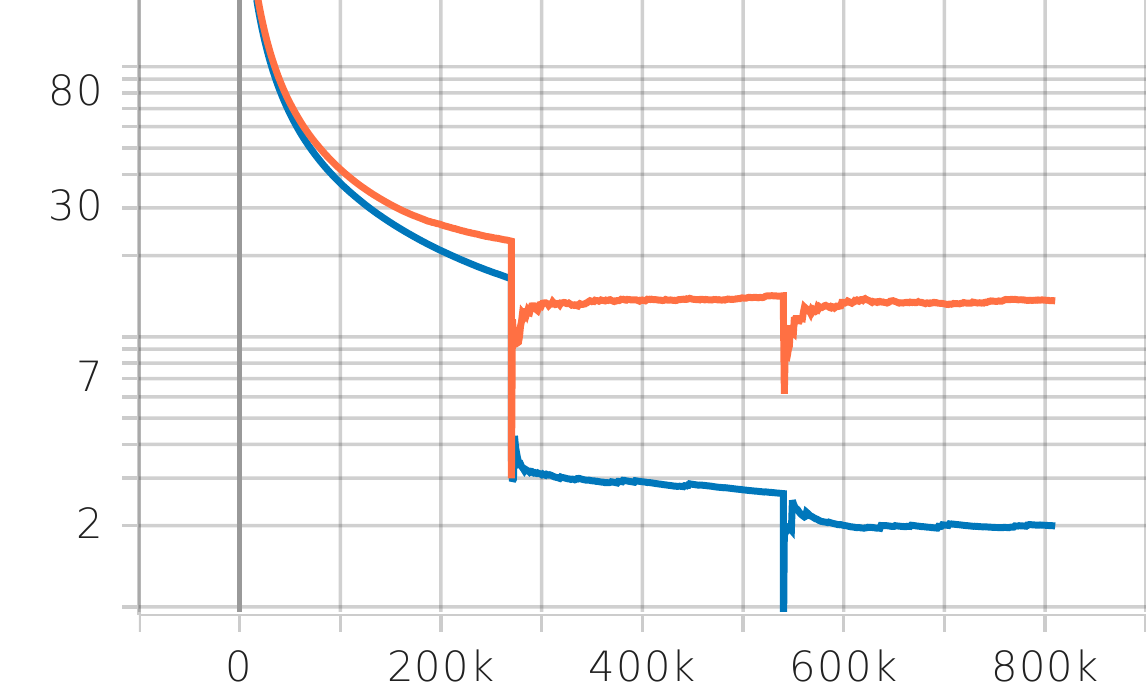}
    \caption{Probabilistic loss computed by Eq. \eqref{eq:nllh}}
  \end{subfigure}
  \hfill
  \begin{subfigure}[b]{0.49\textwidth}
    \centering
    \includegraphics[width=\linewidth]{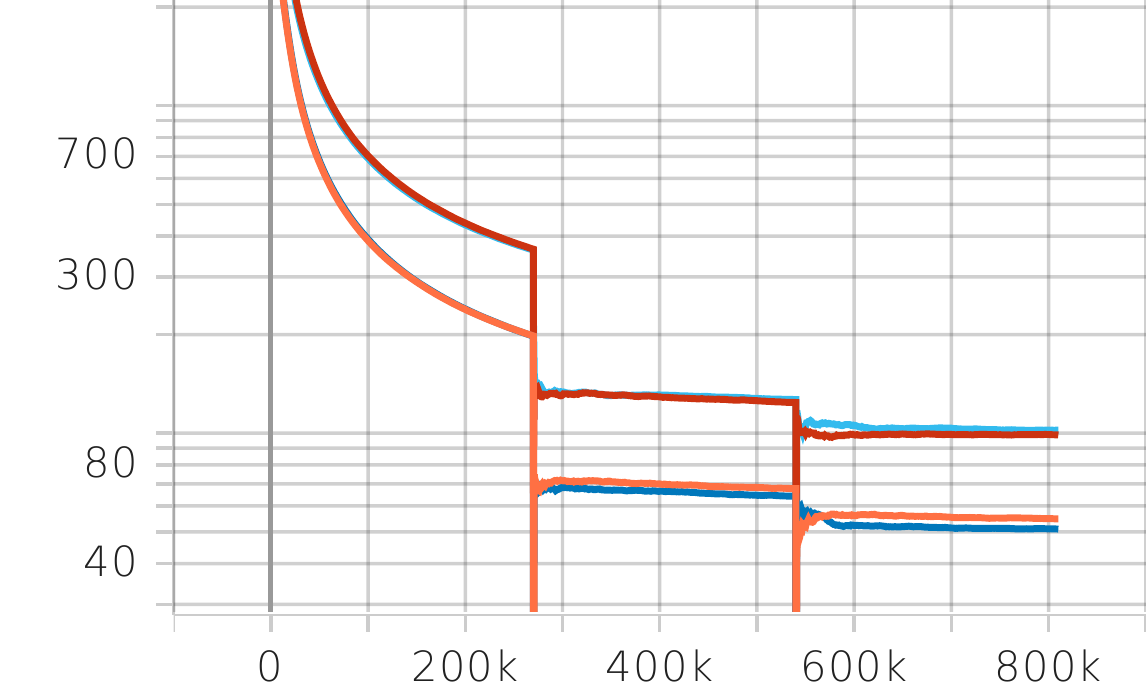}
    \caption{Total loss computed as average}
  \end{subfigure}

  \caption{Loss logged during the training of the proposed models with single and multiple (5) outcomes and Geospatial and Probabilistic loss; KF—Key Feature, MF—Minor Feature; model graph lines by color: \textcolor{orange}{PSOP}, \textcolor{blue}{PMOP}, \textcolor{red}{GSOP}, and \textcolor{cyan}{GMOP} where abbreviations refer to the configuration descriptions provided in Table \ref{tab:model-type-outputs}.
  \label{fig:log-loss}}
\end{figure}

One of the problems mentioned earlier was an alternative for the gazetteers—dictionaries of geographical indexes—used in the previous works containing accurate terms (location naming and LIWs) for geographical objects and their correct geographical coordinates. Considering the noisy user-generated data of the typical model inputs, the idea of feeding a consistent knowledge base of geographic objects into the model appeared practical and has been successfully implemented. The highest performance models were trained on multitask solving that offered an option of parallel learning with a total per-batch training loss as shown in Figure \ref{fig:log-loss}(d). The best strategy was split into key \textit{NON-GEO} (always present 'text' and 'user') and minor \textit{GEO-ONLY} ('place' present only in geo-tagged tweets) text features. In practice, every feature had its individual linear regression layer wrapping the base model that differed in the number of prediction outcomes, but matched in the per-feature loss function type (Geospatial/Probabilistic). 

The evaluation was performed using solely the key feature wrapper layer with no regard for the minor feature wrapper layer utilized only during the finetuning. This approach allowed for maintaining the format of the common model input and mapping the official place descriptions to the locations associated with the primary input texts. In terms of the described setup, the best performance results were achieved using the SOP-type minor loss and MOP-type key loss for 3–5 prediction outcomes. 

Experiments have demonstrated that the geospatial loss of the minor feature \textit{GEO-ONLY} shown in Figure \ref{fig:log-loss}(b) was declining much faster than the error distance of key feature \textit{NON-GEO} shown in Figure \ref{fig:log-loss}(a) since the minor feature had less noisy and more persistent text data associated with the specific geolocations. Such behavior is expected in the case of the relatively high quality of metadata associated with the text which is generally a simpler problem than that of disambiguation within text documents in general \cite{jones2009geographic}.

The straightforward concatenation of the text content and metadata context of the 'place' field would require undesirable randomization of all words/parts in the string before tokenization such that models would not learn to only pay attention to the optional part ('place' metadata) of the input sequence. This statement is supported by the results in Table \ref{tab:worldwide-metrics} which show that the model trained on \textit{ALL} (union of 'text', 'user', and 'place'), as the key feature has underperformed the model trained on the data split to key and minor features. Moreover, the model trained solely on key \textit{NON-GEO} feature with no regard to the place context demonstrated even higher spatial errors during the evaluation. Thus the proposed methodology of multitask learning has demonstrated a significant improvement in the accuracy of the predictions and should be utilized if it's possible. 

Another proposed novelty was a lower-bound limitation of the covariance parameter $\sigma_{\widehat{c}}$ in the GMM output of the probabilistic models. Since the probabilistic loss function computed the negative log-likelihood of the labeled location to fit in the predicted Gaussian distributions, it depended on the Probability Density Function \eqref{eq:pdf}. In the case, $\sigma_{\widehat{c}}$ was approaching its minimum at 0, PDF exceeded 1 driving the loss function to the negative values as shown in Figure \ref{fig:loss-graph-prob}. In terms of probabilistic models, the total loss depended on both spatial and probabilistic components, therefore the latter gained a bigger momentum. As a result, the model paid less attention to the squared Euclidean error distance described in Eq. \eqref{eq:sed} which negatively affected the geospatial accuracy. The suggested solution limited PDF to [0, 1] interval by setting the minimum of $\sigma_{\widehat{c}}$ to $\frac{1}{2\pi}$ such that Probabilistic loss shown in Figure \ref{fig:log-loss}(c) stays in the positive domain. Since the covariance parameter measures model uncertainty about the predicted location point, in the best-case scenarios, the probabilistic loss would depend mainly on the distance error as described in Section \ref{sec:loss-functions}.

According to \cite{iso2017density}, the user-level geolocation is more effective in some cases, however, fine-grained analysis is possible only for the tweet-level predictions and thus the latter was chosen as the main subject of this study. Nonetheless, the problem of the user's home location prediction was explored as one of the possible estimates based on the PMOP-type model predictions. Although this wasn't the foremost performance measure, the results reveal a higher spatial accuracy on the per-user task than on the per-tweet measurements. Hence PMOP-type models are suggested for the user's home location prediction as the comparable transformers-based solution. 

The prediction of tweets and users' home locations on Twitter raises significant privacy concerns. While it has useful applications, it poses risks to privacy. Issues include compromising location privacy, enabling user tracking, exposing personal safety risks, inferring sensitive information, and potential misuse of aggregated data. 

After the training performed on our Twitter archive data, the proposed model was able to predict the location of any Twitter user by processing the publicly available information posted online. While profile information provided by Twitter users is generally considered public, the combination or analysis of such data can have privacy implications. Tweets, being public by default, can also be used for analysis and publication. However, it's worth noting that even seemingly innocuous information, when combined with other data points or analyzed in aggregate, can potentially reveal more about a person's identity or habits. Therefore, while individual profile details may not be inherently sensitive, their combination or analysis might raise privacy concerns.

To address these privacy concerns, it is crucial for researchers, to anonymize or aggregate the data to minimize the risk of re-identification. The data utilized in this study has been protected with strict measures to prevent unauthorized access or breaches. Thus any collected Twitter dataset examples, containing metadata of user-profiles and geo-tags obtained from parsing the tweets, won't be shared due to the potential privacy issues. By balancing the potential benefits of geolocation prediction with the protection of users' privacy, it is possible to mitigate the privacy risks and promote responsible use of location data in Twitter research. 

As for the real-time user location monitoring, Yuan in the work \cite{yuan2013and} focused on tracking the movement of individuals or groups over time. Monitoring user geolocation in time can provide valuable insights for multiple applications such as disease outbreak trackers, urban mobility pattern analyzers, and location-based provider services. Such an analytical framework could be built on top of the proposed models, yet, in this study, only the commonly researched task of the user's home location prediction was properly explored so far.

In addition, there are several different world models that can be utilized for prediction output. While the Plate carrée projection is commonly used due to the Twitter geolocation stored in the WGS84 format, other projections such as the Mercator projection, Robinson projection, conic projection, and Winkel-Tripel projection should be considered as possible alternatives in future studies. 

Besides the BERT variations, the proposed machine learning techniques have the potential to be applied to other base models. In the realm of NLP, various state-of-the-art models have been developed to enhance text analysis tasks, particularly text classification that provides an output representation for the examined regression task. However, this would require adapting the implementation of the wrapper layer to accommodate the shape of the pooler output vector specific to the chosen model. Currently, the approach is designed for linear regression logic, transforming the vector of size 768, which is common to all BERT-based models, into a predefined number of continuous numerical values. Furthermore, in order to ensure suitability for the downstream task of sentence classification, it is important that the base models possess strong contextual understanding and the ability to capture semantic relationships within sentences. 

While acknowledging the availability of newer models, we carefully considered the trade-offs and ultimately selected the BERT model, which was widely adopted and extensively studied for text classification tasks, to strike a balance between computational efficiency and our analytical requirements. Apart from the BERT, RoBERTa (Robustly Optimized BERT) \cite{liu2019roberta}, DistilBERT (Distilled BERT) \cite{sanh2019distilbert}, ALBERT (A Lite BERT) \cite{lan2019albert}, ELECTRA (Efficiently Learning an Encoder that Classifies Token Replacements Accurately) \cite{clark2020electra}, XLNet (eXtreme Language understanding Network) \cite{yang2019xlnet}, and GPT-3 (Generative Pre-trained Transformer 3) \cite{brown2020language} are prominent models in this field that can be finetuned to predict location from the text. 

In the downstream task of text classification, all these models can be finetuned to perform well, although computational resource requirements differ among them. XLNet, BERT, and RoBERTa demand substantial resources due to their size and training needs resulting in a strong performance in text classification tasks. DistilBERT, ALBERT, and ELECTRA offer more efficient alternatives to BERT with similar performance and reduced resource usage. However, compared to the relatively lightweight BERT model, large language models like GPT are computationally more demanding and necessitate significant resources. Moreover, unlike GPT models, which are primarily used for text generation, the BERT model is well-suited for text analysis. 

Overall, location-based sentiment analysis is an important tool for understanding public opinion, social dynamics and patterns, helping decision-makers and researchers to make data-driven decisions, and support the work of various sectors, from marketing to governance. This study provides the NLP-based approach for the estimation of geolocation by processing short text corpora such as social media posts on Twitter. The proposed solution utilizes multitask learning (key and minor features), context data (user and place metadata), and probabilistic output (GMM) to achieve higher spatial accuracy on the tasks of the tweet and the user's home location prediction. Thus contributing to the field of big data analysis with a flexible in geographical granularity setup of the custom BERT-based models. 

\section{Acknowledgments}\label{sec:acknowledge}

The authors acknowledge the Institute of Science and Technology (ISTA) for their material support and for granting access to the Twitter database archive, which was essential for the research.

\bibliographystyle{josisacm}
\bibliography{references}

\begin{thebibliography}{10}

\bibitem{arafat2020demographic}
{\sc Arafat, T.~A., Budi, I., Mahendra, R., and Salehah, D.~A.}
\newblock Demographic analysis of candidates supporter in twitter during
  indonesian presidential election 2019.
\newblock In {\em 2020 International Conference on ICT for Smart Society
  (ICISS)\/} (2020), IEEE, p.~1–6.

\bibitem{bakerman2018twitter}
{\sc Bakerman, J., Pazdernik, K., Wilson, A., Fairchild, G., and Bahran, R.}
\newblock Twitter geolocation: A hybrid approach.
\newblock {\em ACM Transactions on Knowledge Discovery from Data (TKDD) 12}, 3
  (2018), 1–17.

\bibitem{brown2020language}
{\sc Brown, T., Mann, B., Ryder, N., Subbiah, M., Kaplan, J.~D., Dhariwal, P.,
  Neelakantan, A., Shyam, P., Sastry, G., Askell, A., et~al.}
\newblock Language models are few-shot learners.
\newblock {\em Advances in neural information processing systems 33\/} (2020),
  1877–1901.

\bibitem{cao2015inferring}
{\sc Cao, B., Chen, F., Joshi, D., and Philip, S.~Y.}
\newblock Inferring crowd-sourced venues for tweets.
\newblock In {\em 2015 IEEE International Conference on Big Data (Big Data)\/}
  (2015), IEEE, p.~639–648.

\bibitem{cheng2010you}
{\sc Cheng, Z., Caverlee, J., and Lee, K.}
\newblock You are where you tweet: a content-based approach to geo-locating
  twitter users.
\newblock In {\em Proc.~ the 19th ACM international conference on Information
  and knowledge management\/} (2010), p.~759–768.

\bibitem{clark2020electra}
{\sc Clark, K., Luong, M.-T., Le, Q.~V., and Manning, C.~D.}
\newblock Electra: Pre-training text encoders as discriminators rather than
  generators.
\newblock {\em arXiv preprint arXiv:2003.10555\/} (2020).

\bibitem{galal2016enabling}
{\sc Galal, A., and Elkorany, A.}
\newblock Enabling semantic user context to enhance twitter location
  prediction.
\newblock In {\em ICAART (1)\/} (2016), p.~223–230.

\bibitem{han2014text}
{\sc Han, B., Cook, P., and Baldwin, T.}
\newblock Text-based twitter user geolocation prediction.
\newblock {\em Journal of Artificial Intelligence Research 49\/} (2014),
  451–500.

\bibitem{huang2019hierarchical}
{\sc Huang, B., and Carley, K.~M.}
\newblock A hierarchical location prediction neural network for twitter user
  geolocation.
\newblock {\em arXiv preprint arXiv:1910.12941\/} (2019).

\bibitem{hulden2015kernel}
{\sc Hulden, M., Silfverberg, M., and Francom, J.}
\newblock Kernel density estimation for text-based geolocation.
\newblock In {\em Proc.~ the AAAI conference on artificial intelligence\/}
  (2015), vol.~29.

\bibitem{iso2017density}
{\sc Iso, H., Wakamiya, S., and Aramaki, E.}
\newblock Density estimation for geolocation via convolutional mixture density
  network.
\newblock {\em arXiv preprint arXiv:1705.02750\/} (2017).

\bibitem{jones2009geographic}
{\sc Jones, C.~B., Purves, R.~S., Liu, L., and {\"O}zsu, M.~T.}
\newblock Geographic information retrieval.
\newblock {\em Springer reference\/} (2009), 1227–1231.

\bibitem{kinsella2011m}
{\sc Kinsella, S., Murdock, V., and O'Hare, N.}
\newblock " i'm eating a sandwich in glasgow" modeling locations with tweets.
\newblock In {\em Proc.~ the 3rd international workshop on Search and mining
  user-generated contents\/} (2011), p.~61–68.

\bibitem{lan2019albert}
{\sc Lan, Z., Chen, M., Goodman, S., Gimpel, K., Sharma, P., and Soricut, R.}
\newblock Albert: A lite bert for self-supervised learning of language
  representations.
\newblock {\em arXiv preprint arXiv:1909.11942\/} (2019).

\bibitem{li2022transformer}
{\sc Li, M., Lim, K.~H., Guo, T., and Liu, J.}
\newblock A transformer-based framework for poi-level social post geolocation.
\newblock {\em arXiv preprint arXiv:2211.01336\/} (2022).

\bibitem{li2019geoattn}
{\sc Li, S., Zhang, C., Lei, D., Li, J., and Han, J.}
\newblock Geoattn: Localization of social media messages via attentional memory
  network.
\newblock In {\em Proc.~ the 2019 SIAM International Conference on Data
  Mining\/} (2019), SIAM, p.~64–72.

\bibitem{liu2019roberta}
{\sc Liu, Y., Ott, M., Goyal, N., Du, J., Joshi, M., Chen, D., Levy, O., Lewis,
  M., Zettlemoyer, L., and Stoyanov, V.}
\newblock Roberta: A robustly optimized bert pretraining approach.
\newblock {\em arXiv preprint arXiv:1907.11692\/} (2019).

\bibitem{mahmud2014home}
{\sc Mahmud, J., Nichols, J., and Drews, C.}
\newblock Home location identification of twitter users.
\newblock {\em arXiv preprint arXiv:1403.2345\/} (2014).

\bibitem{mcgee2011geographic}
{\sc McGee, J., Caverlee, J.~A., and Cheng, Z.}
\newblock A geographic study of tie strength in social media.
\newblock In {\em Proc.~ the 20th ACM international conference on Information
  and knowledge management\/} (2011), p.~2333–2336.

\bibitem{mcpherson2001birds}
{\sc McPherson, M., Smith-Lovin, L., and Cook, J.~M.}
\newblock Birds of a feather: Homophily in social networks.
\newblock {\em Annual review of sociology 27}, 1 (2001), 415–444.

\bibitem{miura2017unifying}
{\sc Miura, Y., Taniguchi, M., Taniguchi, T., and Ohkuma, T.}
\newblock Unifying text, metadata, and user network representations with a
  neural network for geolocation prediction.
\newblock In {\em Proc.~ the 55th Annual Meeting of the Association for
  Computational Linguistics (Volume 1: Long Papers)\/} (2017), p.~1260–1272.

\bibitem{miyazaki2018twitter}
{\sc Miyazaki, T., Rahimi, A., Cohn, T., and Baldwin, T.}
\newblock Twitter geolocation using knowledge-based methods.
\newblock In {\em Proc.~ the 2018 EMNLP Workshop W-NUT: The 4th Workshop on
  Noisy User-generated Text\/} (2018), p.~7–16.

\bibitem{priedhorsky2014inferring}
{\sc Priedhorsky, R., Culotta, A., and Del~Valle, S.~Y.}
\newblock Inferring the origin locations of tweets with quantitative
  confidence.
\newblock In {\em Proc.~ the 17th ACM conference on Computer supported
  cooperative work \& social computing\/} (2014), p.~1523–1536.

\bibitem{rahimi2017continuous}
{\sc Rahimi, A., Baldwin, T., and Cohn, T.}
\newblock Continuous representation of location for geolocation and lexical
  dialectology using mixture density networks.
\newblock {\em arXiv preprint arXiv:1708.04358\/} (2017).

\bibitem{rahimi2017neural}
{\sc Rahimi, A., Cohn, T., and Baldwin, T.}
\newblock A neural model for user geolocation and lexical dialectology.
\newblock {\em arXiv preprint arXiv:1704.04008\/} (2017).

\bibitem{rahimi2015exploiting}
{\sc Rahimi, A., Vu, D., Cohn, T., and Baldwin, T.}
\newblock Exploiting text and network context for geolocation of social media
  users.
\newblock {\em arXiv preprint arXiv:1506.04803\/} (2015).

\bibitem{roller2012supervised}
{\sc Roller, S., Speriosu, M., Rallapalli, S., Wing, B., and Baldridge, J.}
\newblock Supervised text-based geolocation using language models on an
  adaptive grid.
\newblock In {\em Proc.~ the 2012 joint conference on empirical methods in
  natural language processing and computational natural language learning\/}
  (2012), p.~1500–1510.

\bibitem{sadilek2012finding}
{\sc Sadilek, A., Kautz, H., and Bigham, J.~P.}
\newblock Finding your friends and following them to where you are.
\newblock In {\em Proc.~ the fifth ACM international conference on Web search
  and data mining\/} (2012), p.~723–732.

\bibitem{sanh2019distilbert}
{\sc Sanh, V., Debut, L., Chaumond, J., and Wolf, T.}
\newblock Distilbert, a distilled version of bert: smaller, faster, cheaper and
  lighter.
\newblock {\em arXiv preprint arXiv:1910.01108\/} (2019).

\bibitem{scherrer2021social}
{\sc Scherrer, Y., and Ljube{\v{s}}i{\'c}, N.}
\newblock Social media variety geolocation with geobert.
\newblock In {\em Proc.~ the Eighth Workshop on NLP for Similar Languages,
  Varieties and Dialects\/} (2021), The Association for Computational
  Linguistics.

\bibitem{schulz2013multi}
{\sc Schulz, A., Hadjakos, A., Paulheim, H., Nachtwey, J., and
  M{\"u}hlh{\"a}user, M.}
\newblock A multi-indicator approach for geolocalization of tweets.
\newblock In {\em Proc.~ the International AAAI Conference on Web and Social
  Media\/} (2013), vol.~7, p.~573–582.

\bibitem{simanjuntak2022we}
{\sc Simanjuntak, L.~F., Mahendra, R., and Yulianti, E.}
\newblock We know you are living in bali: Location prediction of twitter users
  using bert language model.
\newblock {\em Big Data and Cognitive Computing 6}, 3 (2022), 77.

\bibitem{villegas2020point}
{\sc Villegas, D.~S., Preo{\c{t}}iuc-Pietro, D., and Aletras, N.}
\newblock Point-of-interest type inference from social media text.
\newblock {\em arXiv preprint arXiv:2009.14734\/} (2020).

\bibitem{wakamiya2018twitter}
{\sc Wakamiya, S., Kawai, Y., Aramaki, E., et~al.}
\newblock Twitter-based influenza detection after flu peak via tweets with
  indirect information: text mining study.
\newblock {\em JMIR public health and surveillance 4}, 3 (2018), e8627.

\bibitem{wing2011simple}
{\sc Wing, B., and Baldridge, J.}
\newblock Simple supervised document geolocation with geodesic grids.
\newblock In {\em Proc.~ the 49th annual meeting of the association for
  computational linguistics: Human language technologies\/} (2011),
  p.~955–964.

\bibitem{wing2014hierarchical}
{\sc Wing, B., and Baldridge, J.}
\newblock Hierarchical discriminative classification for text-based
  geolocation.
\newblock In {\em Proc.~ the 2014 conference on empirical methods in natural
  language processing (EMNLP)\/} (2014), p.~336–348.

\bibitem{yang2019xlnet}
{\sc Yang, Z., Dai, Z., Yang, Y., Carbonell, J., Salakhutdinov, R.~R., and Le,
  Q.~V.}
\newblock Xlnet: Generalized autoregressive pretraining for language
  understanding.
\newblock {\em Advances in neural information processing systems 32\/} (2019).

\bibitem{yaqub2020location}
{\sc Yaqub, U., Sharma, N., Pabreja, R., Chun, S.~A., Atluri, V., and Vaidya,
  J.}
\newblock Location-based sentiment analyses and visualization of twitter
  election data.
\newblock {\em Digital Government: Research and Practice 1}, 2 (2020), 1–19.

\bibitem{yuan2013and}
{\sc Yuan, Q., Cong, G., Ma, Z., Sun, A., and Thalmann, N.~M.}
\newblock Who, where, when and what: discover spatio-temporal topics for
  twitter users.
\newblock In {\em Proc.~ the 19th ACM SIGKDD international conference on
  Knowledge discovery and data mining\/} (2013), p.~605–613.

\bibitem{zhai2017study}
{\sc Zhai, C., and Lafferty, J.}
\newblock A study of smoothing methods for language models applied to ad hoc
  information retrieval.
\newblock In {\em ACM SIGIR Forum\/} (2017), vol.~51, ACM New York, NY, USA,
  p.~268–276.

\bibitem{zhang2014geocoding}
{\sc Zhang, W., and Gelernter, J.}
\newblock Geocoding location expressions in twitter messages: A preference
  learning method.
\newblock {\em Journal of Spatial Information Science}, 9 (2014), 37–70.

\bibitem{zheng2020social}
{\sc Zheng, C., Jiang, J.-Y., Zhou, Y., Young, S.~D., and Wang, W.}
\newblock Social media user geolocation via hybrid attention.
\newblock In {\em Proc.~ the 43rd International ACM SIGIR Conference on
  Research and Development in Information Retrieval\/} (2020), p.~1641–1644.

\bibitem{zheng2018survey}
{\sc Zheng, X., Han, J., and Sun, A.}
\newblock A survey of location prediction on twitter.
\newblock {\em IEEE Transactions on Knowledge and Data Engineering 30}, 9
  (2018), 1652–1671.

\bibitem{zhong2020multiple}
{\sc Zhong, T., Wang, T., Wang, J., Wu, J., and Zhou, F.}
\newblock Multiple-aspect attentional graph neural networks for online social
  network user localization.
\newblock {\em IEEE Access 8\/} (2020), 95223–95234.

\bibitem{ZHOU20221}
{\sc Zhou, F., Wang, T., Zhong, T., and Trajcevski, G.}
\newblock Identifying user geolocation with hierarchical graph neural networks
  and explainable fusion.
\newblock {\em Information Fusion 81\/} (2022), 1--13.
\newblock https://doi.org/10.1016/j.inffus.2021.11.004.

\end{thebibliography}

\appendix

\section{Per-user geolocation estimation}\label{sec:per-user-predict}

In this work, the focus was primarily on solving the problem of tweet geolocation prediction, thus there is no direct way to predict user's home location using the proposed approaches. However, the PMOP-type models were leveraged to obtain a set of per-tweet predictions for a single user which was then used to estimate the most probable user's home locations in the form of GMOP-type output. Note that in this case the number of selected location points could vary among users, and their weights were set manually based on the summary scores computed for each outcome. 

The summarizing of PMOP-type output of $M$ outcomes was computed on the grid of $S \cdot M$ GMM peaks gathered from the $S$ user's per-tweet predictions put among points of the ground grid $G$ generated on a Plate carrée projection map with step 10:

\[\mathbf{T} = \{\boldsymbol{\widehat{\mu}}_{i,j} \mid 1 \leq i \leq S, 1 \leq j \leq M\}; \quad\boldsymbol{\widehat{\mu}}_{i,j}=(\widehat{y}_{lon}, \widehat{y}_{lat}); \quad \mathbf{T} \in \mathbb{R}^{(S \cdot M) \times (S \cdot M)}\]

\[\mathbf{C} = \mathbf{G} \cup \mathbf{T}; \quad \mathbf{C}_{i,j} = (y_{lon, i}, y_{lat, j}); \quad \mathbf{G} \in \mathbb{R}^{36 \times 19}\]

The union $\mathbf{C}$ provided a multi-set of all GMM peaks $\mathbf{T}$ merged with a background of grid points $\mathbf{G}$. The average of all $S$ per-tweet scores was computed for each grid point $\mathbf{C}_{i,j}$ as its likelihood to fit into a predicted GMM of $M$ peaks defined by their weights $W$, two-dimensional means $\boldsymbol{\mu}$, and covariance matrices $\mathbf{\Sigma}$:

\[summary(\mathbf{C}_{i,j}) = \frac{1}{S} \sum_{s=1}^{S} \sum_{m=1}^{M} W_{s,m} \cdot N(\mathbf{C}_{i,j} \mid \boldsymbol{\widehat{\mu}}_{s,m}, \mathbf{\Sigma}_{s,m}); \quad\mathbf{C}_{i,j} \in \mathbf{C}\]

Thus forming a two-dimensional matrix $\mathbf{Z}$ containing the average of $S$ probabilities as values of the summarizing function for all grid points $\mathbf{C}_{i,j}$ such that:

\[Z(i, j) = summary(\mathbf{C}_{i,j}); \quad summary: \mathbf{R}^2 \rightarrow \mathbf{R}^2\]

Therefore, $Z(i,j)$ was the function value (summary score) of a two-dimensional matrix at point $(i,j)$, and $M_fZ(i,j)$ was the filtered function value of $Z(i,j)$ using a 10 by 10 maxima filter footprint. The formula took the maximum value of $\mathbf{Z}$ within $10 \times 10$ window and assigned it to $M_{f}Z(i,j)$. This operation could be repeated for every point $(i,j)$ in the matrix $\mathbf{Z}$ to obtain the filtered matrix $M_fZ$:

\[M_fZ(i,j) = max_{p=-4}^{5} max_{q=-4}^{5} Z(i+p,j+q)\]

Then the set of local maxima of $\mathbf{Z}$ was defined as:

\[ \mathbf{L}^{user} = \{ (i,j) \mid (\mathbf{Z}(i,j) = M_fZ(i,j)) \land (i, j) \in \mathbf{C} \}\]

where $\mathbf{C}$ was the multi-set of grid points, and $(i,j) \in \mathbf{C}$ means that $(i,j)$ was a point in the grid $\mathbf{C}$. 

Note that the bag of non-unique points $\mathbf{C}$ was reduced to the set of unique local maxima points $\mathbf{L}^{user}$ that could consist of a single coordinate pair or multiple unweighted ones. Assuming that $\mathbf{L}^{user}$ contained multiple location points, the top $K$ most probable user's home locations were obtained as the first $K$ elements of the sorted in descending order summary scores for all points in $\mathbf{L}^{user}$ which is referenced as $\mathbf{Z}^{top-K}$:  

\[ \mathbf{L}^{top-K} = \{ (i,j) \mid (\mathbf{Z}(i,j) \in \mathbf{Z}^{top-K} \land (i, j) \in \mathbf{L}_{user} \}; \quad \mathbf{L}^{top-K} \in \mathbb{R}^{K \times 2} \]

Then, the coordinate pairs assigned to location indices $(i, j)$ in the set of estimated locations $\mathbf{L}^{top-K}$ were obtained from the initial grid $\mathbf{C}$ as follows:

\[\mathbf{Y}_{k}^{user} = \mathbf{C}(\mathbf{L}_{k}^{top-K}) = (y_{lon, i}, y_{lat, j}); \quad \mathbf{Y}^{user} \in \mathbb{R}^{K \times 2}\]

Moreover, weights of the selected location points in $\mathbf{Y}^{user}$ were estimated by the application of Eq. \eqref{eq:sm} to their corresponding summary scores in $\mathbf{Z}^{top-K}$ as follows: 

\[W_k^{user}=\frac{e^{\mathbf{Z}_{k}^{top-K}}}{\sum_{j=1}^{K}e^{\mathbf{Z}_{j}^{top-K}}}; \quad \sum_{k=1}^{K}W_k^{user} = 1; \quad W_k^{user} \in [0, 1]\]

Although the initial model output format had to be of PMOP-type, the estimated points $\mathbf{Y}^{user}$ and their weights $W^{user}$ formed a result similar to the GMOP-type output. Note that the set of multiple prediction outcomes for the task of user's home location prediction could be formed only in the case $\mathbf{L}^{user}$ had more than one unique location. Otherwise, the estimation would result in a GSOP-type output containing a single coordinate pair $Y^{user}$ which excludes the calculation of $W^{user}$.    

\section{Performance metrics} \label{sec:performance-metrics}

\subsection{Geospatial metrics}\label{sec:spat-metrics}

\begin{itemize}

\item SAE—\textit{simple accuracy error} metric is utilized to measure the geographic distance between two points on the Earth's surface. This is achieved through the application of the Haversine formula for calculating the great-circle distance in kilometers. The formula is as follows:

\[SAE_{SOP}=Hav(\mathbf{Y}, \mathbf{\widehat{Y}})=D_{H}\]

where $\mathbf{Y}$ and $\mathbf{\widehat{Y}}$ represent the original and predicted points, respectively. When calculating the MOP-type model metrics, the SAE is computed as the weighted linear combination of all $M$ outcomes: 

\[SAE_{MOP}=\sum_{i=1}^{M}W_{i}Hav(\mathbf{Y}, \mathbf{\widehat{Y}}_{i}) = D_{H}\]

The mean and median of the SAE for the validation dataset are utilized as key performance indicators. It is worth noting that during the finetuning process, the calculation of the spatial error is simplified to the Euclidean distance formula.

\item Acc@161—the percentage of predicted locations that are within a 161km (100 miles) radius of the actual location. 

\[Acc@161 = \frac{\left|{i: D_{H,i} \leq 161 }\right|}{N}\cdot 100\]

where $D_H$ is a set of $N$ elements representing the great-circle distance between the predicted locations and the labeled locations, and $\left|{i: D_{H,i} \leq 161 }\right|$ represents the number of elements in $D_H$ that are less than or equal to 161 km. 

Note that this metric based on the imperial units was used as one of the most popular in the area of location prediction due to the great contribution of researchers from the United States.   
\end{itemize}

\subsection{Probabilistic metrics}\label{sec:prob-metrics}

\begin{itemize}

\item CAE—\textit{comprehensive accuracy error} metric measures the expected distance between the true origin of the tweet and a random point generated from the GMM predicted by the model. The Haversine formula is used to calculate distances from the original point to each Gaussian sub-population of 100 samples. Bakerman in his work \cite{bakerman2018twitter} suggests the Monte Carlo method to compute CAE as the mean of the sub-population distance errors: 

\[CAE_{SOP}\approx \frac{1}{\left| z \right|} \cdot \sum_{\mathbf{\widehat{Y}} \in z} Hav(\mathbf{Y}, \mathbf{\widehat{Y}})\]

where $z$ is a sample from the Gaussian sub-population. The total CAE of the GMM is computed as the weighted linear combination of all $M$ Gaussian peaks:

\[CAE_{MOP}\approx \sum_{i=1}^{M} \frac{W_{i}}{\left| z_i \right|} \cdot \sum_{\mathbf{\widehat{Y}} \in z_i} Hav(\mathbf{Y}, \mathbf{\widehat{Y}})\]

\item $PRA_{\alpha}$—\textit{prediction region area} metric is the area covered by $\alpha \cdot 100\%$ density of the predicted GMM: 

\[PRA_{\alpha, SOP}=\pi\chi_2^2(1-\alpha) \sigma_{\widehat{c}}\]

where $\sigma$ is a determinant of the spherical covariance matrix $\mathbf{\Sigma}$ from the predicted GMM peak $N(\boldsymbol{\widehat{\mu}},\mathbf{\Sigma})$. In the case of MOP, It's computed as the weighted linear combination of such area for each of $M$ Gaussian peaks, which depends mainly on their covariance parameters and weights: 

\[PRA_{\alpha, MOP}=\pi\chi_2^2(1-\alpha) \cdot \sum_{i=1}^{M} W_i \sigma_i\]

\item $COV_{\alpha}$—\textit{coverage} metric measures the proportion of times the prediction region $PRA_{\alpha}$ covers the true origin of the tweet. Following Bakerman's approach, the original geographic point $\mathbf{Y}$ is within the boundary of an ellipse with center $\boldsymbol{\widehat{\mu}}$ and covariance matrix $\mathbf{\Sigma}$ if:

\[\frac{D}{\sigma} \leq \chi_2^2(1-\alpha); \quad D = Euclidean(\mathbf{Y}, \boldsymbol{\widehat{\mu}})\]

Note that in the MOP case, $COV_{\alpha}$ is calculated as the overall mean of $N$ dataset samples with $M$ outcomes. 

\end{itemize}

\section{Loss functions} \label{sec:loss-functions}

\begin{figure}[!htb]
\centering
     \includegraphics[width=1.0\textwidth]{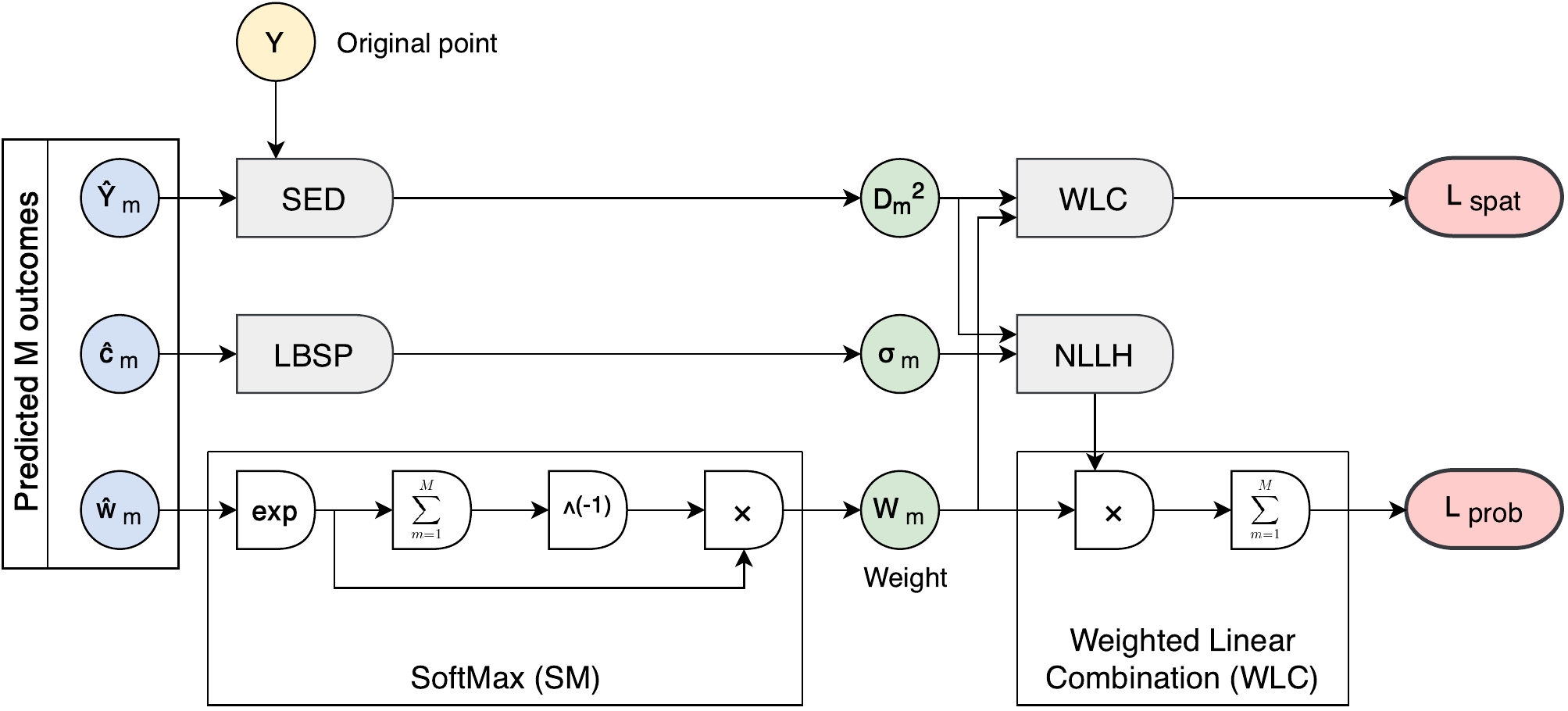}
      \caption{Multiple Outcomes Prediction (MOP) loss functions computational graph including visualization of Weighted Linear Combination \eqref{eq:wlc} and SoftMax (SM) components. \\
       $\widehat{Y}_m$—point, $\widehat{c}_m$—covariance parameter, and $\widehat{w}_m$—weight predicted for a single outcome $m$ \\
       $D_m^2$—squared error distance, $\sigma_m$—adjusted covariance (confidence), and $\widehat{W}_m$—normalized weight (significance) for a single outcome $m$}
       \label{fig:mop-loss}
\end{figure}

% The total loss per text feature $f$ had an essential geospatial component, such that $L_{f} = L_{spat}$, and optionally the probabilistic loss added to it. The per-feature loss for the probabilistic models were calculated in three ways: average or sum of $L_{spat}$ and $L_{prob}$, or $L_{prob}$ with no regard for $L_{spat}$. The first option has slightly outperformed the second option and has significantly outperformed the last option, thus per-feature loss was calculated as follows: 

% \[L_{f} = \frac{L_{spat} + L_{prob}}{2}\]

% Finally, the total loss as the average of a single KF and $K$ MFs was calculated to handle multiple textual features $F$ of a single tweet. Note that the number of outcomes $M$ was related solely to the KF output, while the MFs retained SOP-type outputs. However, the number of MFs $K$ was equal to $F-1$ and could be any integer greater or equal to zero. 

% \[L_{total} = \frac{\sum_{i=1}^{F}L_{i}}{F}; \quad F\geq 1\]

\begin{figure}[!htb]
\centering
     \includegraphics[width=1.0\textwidth]{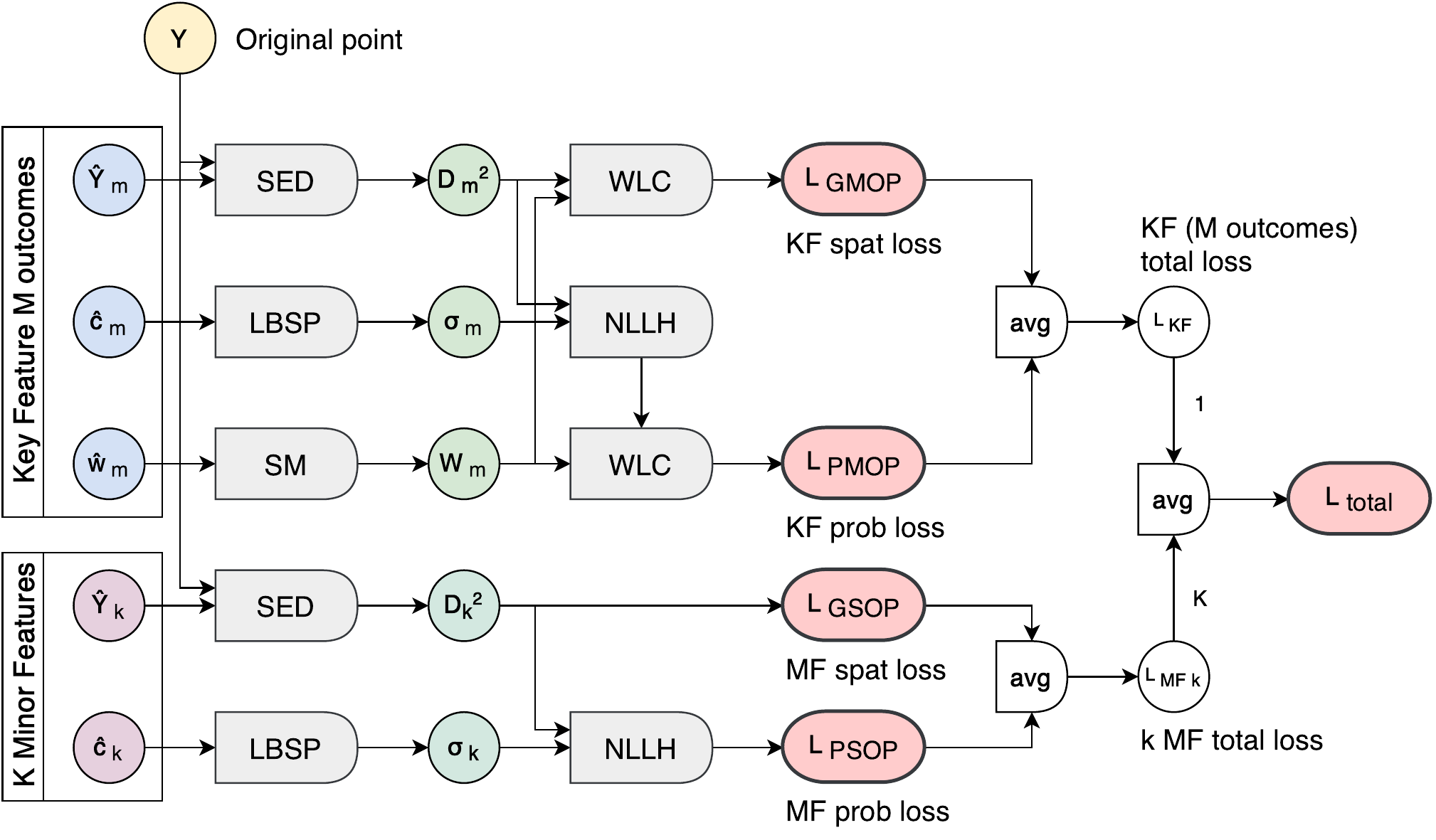}
      \caption{Total loss functions computational graph for the MOP-type Key Feature (KF) with M outcomes, and K Minor Features (MF) of the SOP-type.}
       \label{fig:total-loss}
\end{figure}

% \nocite{*}

\end{document}